\documentclass{article}

\usepackage{microtype}
\usepackage{graphicx}
\usepackage{subcaption}
\usepackage{booktabs} % for professional tables

\usepackage{hyperref}

\usepackage[table]{xcolor}
\usepackage{makecell}
\usepackage{pifont}
\usepackage{enumitem}
\usepackage{amsmath}
\usepackage{multirow}

\usepackage[accepted]{icml2026}

\usepackage{amsmath}
\usepackage{amssymb}
\usepackage{mathtools}
\usepackage{amsthm}

\usepackage[capitalize,noabbrev]{cleveref}

\theoremstyle{plain}

\theoremstyle{definition}

\theoremstyle{remark}

\usepackage[textsize=tiny]{todonotes}

\icmltitlerunning{Redundancy-Aware Diffusion for Multi-Agent Communication Structure Generation}

\begin{document}

\twocolumn[
  %\icmltitle{RADAR: Redundancy-Aware Multi-Agent Communication Topology Design via Graph Diffusion}
  \icmltitle{RADAR: Redundancy-Aware Diffusion for Multi-Agent Communication Structure Generation}



  \icmlsetsymbol{equal}{*}

  \begin{icmlauthorlist}
    \icmlauthor{Zhen Zhang}{nju}
    \icmlauthor{Wanjing Zhou}{nju}
    \icmlauthor{Juncheng Li}{zju}
    \icmlauthor{Hao Fei}{oxford}
    \icmlauthor{Jun Wen}{mbzuai}
    \icmlauthor{Wei Ji}{nju}
  \end{icmlauthorlist}

  \icmlaffiliation{nju}{National Key Laboratory for Novel Software Technology, Nanjing University, China,}
  \icmlaffiliation{zju}{Zhejiang University,}
  \icmlaffiliation{oxford}{University of Oxford,}
  \icmlaffiliation{mbzuai}{Mohamed bin Zayed University of Artificial Intelligence}

  \icmlcorrespondingauthor{Zhen Zhang}{zhen\_zhang@nju.edu.cn}
  \icmlcorrespondingauthor{Wei Ji}{weiji@nju.edu.cn}

  \vskip 0.3in
]



\printAffiliationsAndNotice{}  % no special notice (required even if empty)

\begin{abstract}
  Compared with individual agents, large language model based multi-agent systems have shown great capabilities consistently across diverse tasks, including code generation, mathematical reasoning, and planning, etc. Despite their impressive performance, the effectiveness and robustness of these systems heavily rely on their communication topology, which is often fixed or generated in a single step. This restricts fine-grained structural exploration and flexible composition, resulting in excessive token utilization on simple tasks while limiting capability on complicated tasks. To mitigate this challenge, we introduce RADAR, a redundancy-aware and query-adaptive generative framework that actively reduce communication overhead. Motivated by recent progress in conditional discrete graph diffusion models, we formulate communication topology design as a step-by-step generation process, guided by the effective size of the graph. Comprehensive experiments on six benchmarks demonstrate that RADAR consistently outperforms recent baselines, achieving higher accuracy, lower token consumption, and greater robustness across diverse scenarios. Our code and data are available at \url{https://github.com/cszhangzhen/RADAR}.
\end{abstract}

\section{Introduction}
Large Language Model (LLM) based agents have achieved great success across a wide spectrum of domains, including code generation \cite{zhang2024codeagent}, question answering \cite{xu2024generate} and web navigation \cite{chae2025web}, etc. Beyond single-agent paradigms, prior work shows that multi-agent configurations, whether cooperatively \cite{zhuge2024gptswarm,zhang2025gdesigner} or competitively \cite{zhu2025multiagentbench,wu2024shall}, can surpass the capabilities of individual agents \cite{du2023improving,zhang2025cut}, highlighting the emergence of collective intelligence \cite{piatti2024cooperate,chen2025debatecoder}. However, the emergence of this collective intelligence is primarily shaped by the design of the communication topology, which specifies agent connectivity and information flow \cite{zhang2025multiagent,hu2025automated}. This highlights the crucial of effective collaboration graph design for multi-agent system performance, making it a key focus of research in this area.

Prior work on multi-agent systems has largely been constrained to static, hand-crafted configurations, such as chain structures that impose sequential execution \cite{wei2022chain}, star topologies that enable centralized coordination \cite{jin2025controlling}, tree architectures that support hierarchical collaboration \cite{yao2023tree}, and fully connected topologies that facilitate global communication. Although effective in specific scenarios, these collaboration architectures lack the flexibility required to generalize across diverse tasks. For instance, a simple arithmetic task might require a brief, linear exchange, whereas complex code generation often demands more elaborate collaborative structures. Therefore, applying a single fixed collaboration pattern across all tasks either incurs redundant communication overhead for simple problems or limits performance on more complicate ones \cite{zhang2025multiagent,jiang2025dynamic}.

To enhance flexibility, more recent research has shifted towards automatically designing task-adaptive multi-agent systems \cite{zhang2025gdesigner,li2025assemble}. These methods can be broadly classified into three categories: agentic profiling methods \cite{chen2024autoagents,yuan2025evoagent}, which employs a coordinating agent to facilitate information sharing and environment adaptation; search-based models \cite{zhang2025multiagent,shang2025agentsquare,zhang2025aflow}, which explores the design space of multi-agent configurations, and graph learning approaches \cite{zhuge2024gptswarm,zhang2025gdesigner,wang2025agentdropout}, which learn inter-agent connectivity patterns to support task collaboration. Despite their diverse design paradigms, these approaches are constrained by fundamental limitations. Agentic profiling methods \cite{zhang2025aflow,yuan2025evoagent} rely heavily on a coordinating or meta-agent to profile roles, route information, or adjust collaboration strategies, which suffer from single-point bottlenecks. Search-based models \cite{shang2025agentsquare,zhang2025aflow} explore the design space of multi-agent configurations through heuristic or combinatorial search, which are often computationally expensive and poorly scalable. Graph learning approaches \cite{zhuge2024gptswarm,zhang2025gdesigner} explicitly learn inter-agent connectivity patterns, but they typically generate graphs in a single step conditioned on task descriptions, which restricts their ability to capture fine-grained and adaptive structures. These limitations highlight the need for more principled and flexible approaches that can dynamically generate collaboration topologies, efficiently adapt to diverse tasks, and balance structural expressiveness with communication efficiency.

\begin{figure}
    \centering
    \includegraphics[width=0.95\linewidth]{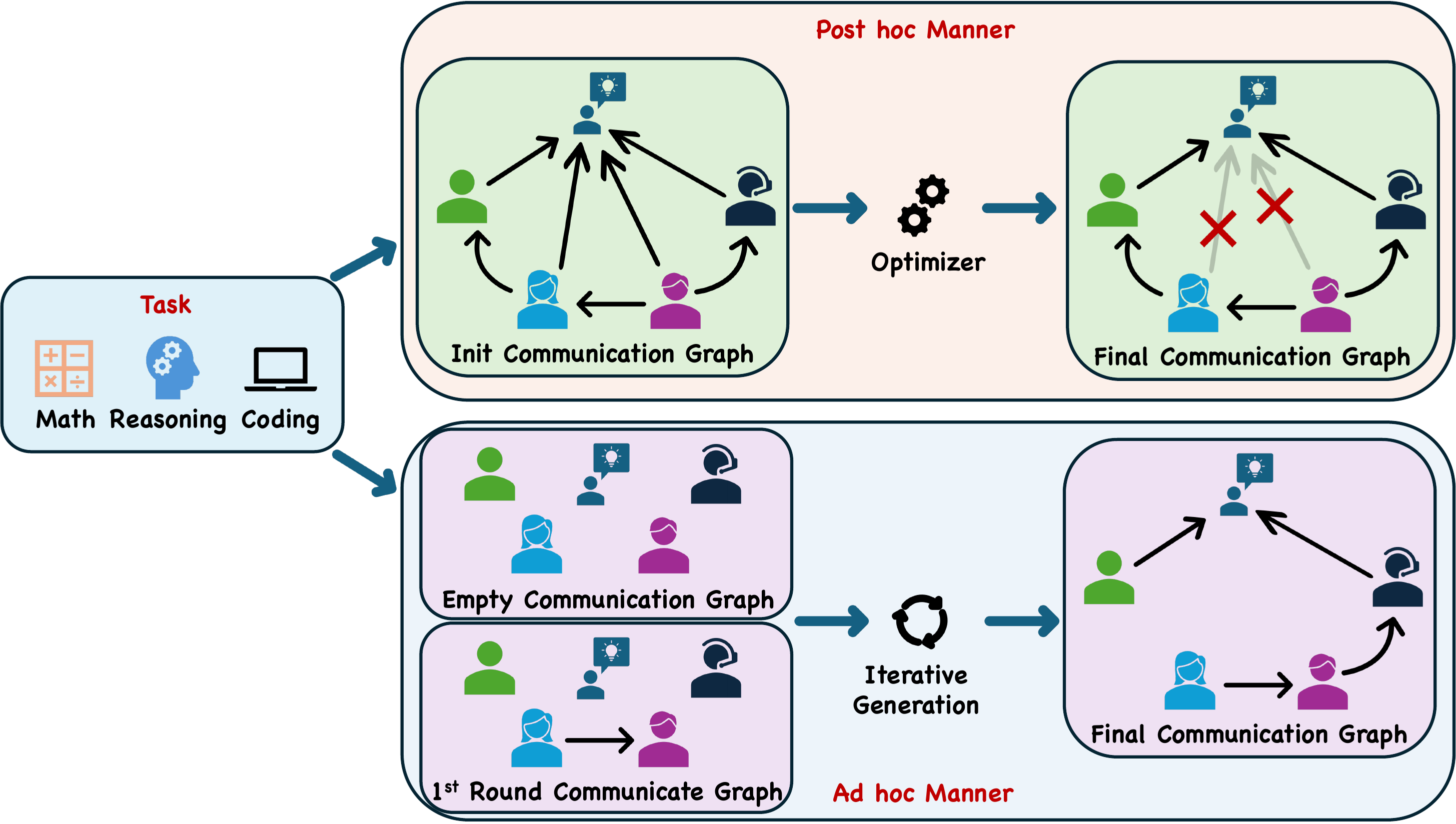}
    \caption{Comparison of workflows for designing communication topologies in LLM multi-agent systems. Compared with an ad hoc design approach, the post hoc paradigm might still incur redundant communication and fail to uncover novel structures.}
    \label{fig:intro}
    \vspace{-0.3in}
\end{figure}

At the same time, as multi-agent systems become more structurally elaborate, a complementary challenge emerges: their communication costs rise significantly due to overly complicated and often redundant collaboration designs. While rich interaction structures offer strong expressiveness, they can introduce unnecessary message passing, excessive token consumption, and coordination overhead, ultimately limiting scalability and efficiency. Empirical results in \cite{zhang2025cut} demonstrate that more complex communication strategies require $2 \sim 11.8\times$ more tokens than simple chain topologies. To further improve collaboration efficiency, Zhang et al. \yrcite{zhang2025cut} focus on pruning redundant connections in communication graphs, whereas Wang et al. \yrcite{wang2025agentdropout} propose to selectively dropping low-contributing agents. However, these methods operate on a predefined collection of agents and largely static interaction structures, where learning is limited to local modifications such as edge pruning or node dropping. Thus, they optimize efficiency in a post hoc manner, rather than jointly reasoning about structure formation and redundancy control during topology construction. As illustrated in Figure \ref{fig:intro}, they lack the capability to fundamentally redesign or generate collaboration topologies from scratch in a communication efficient manner, limiting their expressiveness and adaptability to diverse or evolving task requirements.

To address the above challenges, we propose \underline{\textbf{R}}edundancy-\underline{\textbf{A}}ware \underline{\textbf{D}}iffusion for Multi-\underline{\textbf{A}}gent communication st\underline{\textbf{R}}ucture generation (\textbf{RADAR}), which synthesizes the entire collaboration graph via iterative conditional graph diffusion models \cite{kong2023autoregressive,chen2023efficient}. Specifically, we incorporate the concept of effective size \cite{ronald1992structural}, which measures the non-redundant portion of a node's ego network, into the graph generation process to guide the construction of low-redundancy multi-agent communication structures. Meanwhile, this iterative, step-wise generation paradigm not only supports fine-grained exploration of the design space but also leverage the evolving partial graph at each step, enabling informed structural decisions and progressive mitigation of emerging redundancies. During inference, RADAR incorporates query-dependent contextual conditioning, thereby generating collective intelligence tailored to the specific task. Comprehensive evaluations are performed on six widely used benchmarks spanning reasoning, code generation and mathematical problem solving. The experimental results show that the proposed RADAR demonstrates substantial improvements compared with existing state-of-the-art baselines.

In summary, this work makes the following contributions:
\begin{itemize}[leftmargin=*]
    \item \textbf{Paradigm Redesign:} We develop an iterative, step-wise framework, which enables fine-grained generation of the multi-agent collaboration space, rather than relying on one-step generation. 
    \item \textbf{Practical Framework:} The proposed framework automatically generates high-quality multi-agent collaboration systems and adaptively select efficient and high-performing solutions for queries of varying difficulty.
    \item \textbf{Experimental Evaluation:} Extensive experimental results across six public datasets show that our proposed model surpasses state-of-the-art baselines with different gains, while delivering improved token efficiency and enhanced robustness.
\end{itemize}

\section{Related Works}
\textbf{LLM-agent Collaboration.} The success of single agent systems \cite{shen2023hugginggpt,wang2024rethinking,song2023llm} has inspired the development of multi-agent collaboration, giving rise to emergent collective intelligence. Numerous paradigms have been introduced to facilitate collaboration among multiple agents, including non-interactive query schemes, chain-of-thought prompting to debate mechanisms and fixed tree or star structures. Among them, LLM-Debate \cite{du2023improving} orchestrates multi-round debates in which agents independently articulate and refine their reasoning before converging on a final answer. MetaGPT \cite{hong2024metagpt} formalizes collaboration by encoding standardized operating procedures into structured prompt sequences, yielding a sequential workflow. AutoGen \cite{wu2024autogen} supports flexible, conversational interaction patterns, including nested-chat, tree, and star topologies, enabling customizable coordination among agents. Nonetheless, their reliance on manual designs limits agents' capability to adapt to evolving scenarios.

\textbf{Multi-agents as Graphs.} Graphs provide a powerful and flexible framework for capturing interactions among diverse objects \cite{liu2024revisiting,zhang2025towards}. Communication in multi-agents systems can naturally be represented as a graph, with nodes representing agents and edges encoding the flow of information. Early efforts primarily adopt complete or random graph structures \cite{qian2025scaling}, which assumes dense or arbitrary connection would be sufficient to facilitate effective information exchange among agents. To eliminate manual pipeline construction, more recent research has focused on automated topology learning. For instance, GPTSwarm \cite{zhuge2024gptswarm} incorporates node-level optimization to adapt agent prompts and edge-level optimization to refine communication patterns. G-Designer \cite{zhang2025gdesigner} leverages a variational graph auto-encoder for encoding and decoding multi-agent interaction topologies. MaAS \cite{zhang2025multiagent} learns a probabilistic, continuous architecture distribution from which task-adaptive multi-agent systems can be sampled for different queries. However, these approaches remain limited by their one-step generation mechanisms, which prevent them from exploring fine-grained collaboration structures and capturing the progressive dependencies inherent in complex multi-agent coordination.

\textbf{Graph Diffusion Models.} 
The rise of generative modeling has unlocked powerful mechanisms for graph synthesis, with conditional graph diffusion models achieving notable success in domains such as protein design \cite{yi2023graph}, molecular generation \cite{liu2024graph}, materials \cite{klipfel2024vector} etc. Inspired by these advances, recent multi-agent systems research has begun to explore generating collaboration structures from scratch. For example, ARG-Designer \cite{li2025assemble} constructs collaboration graphs using autoregressive modeling, while GTD \cite{jiang2025dynamic} employs conditional discrete graph diffusion for dynamic topology generation. Nevertheless, existing methods fail to explicitly reason about redundancy during structure formation, leading to unnecessary token overhead for simple tasks or limited effectiveness on more challenging problems. To overcome this limitation, we devise a novel redundancy-aware graph generation framework that conditions on effective topology size and constructs communication graphs step by step, enabling more adaptive and robust multi-agent collaboration.

\section{Problem Formulation}
In this section, we begin by establishing the notations and formalizing the key concepts from a topological perspective, followed by a  formal definition of the LLM-based multi-agent communication protocol design.

\subsection{Topological Structure}
The multi-agent system can be naturally modeled as a directed graph $\mathcal{G} = (\mathcal{V}, \mathcal{E})$ with node set $\mathcal{V}$ and edge set $\mathcal{E}$. Each agent $v_i \in \mathcal{V}$ is formally defined as:
\begin{equation}
    v_i = \{\texttt{Base}_{i}, \texttt{Role}_{i}, \texttt{State}_{i}, \texttt{Plugin}_{i} \},
\end{equation}
where \texttt{Base}$_i$ refers to the specific instance of a large language model; \texttt{Role}$_{i}$ denotes the functional role assigned to agent $i$; \texttt{State}$_{i}$ encapsulates the agent's accumulated knowledge and prior interactions; \texttt{Plugin}$_{i}$ specifies the external tools and plugins available to the agent, including web searcher, code compiler or file readers, etc \cite{wu2025agentic,wolflein2025llm}. Upon receiving the prompt $\mathcal{P}_i$, each agent $v_i$ produces the corresponding response $\mathcal{R}_{i}$:
\begin{equation}
    \mathcal{R}_i = v_i(\mathcal{P}_i) = v_i(\mathcal{P}_{\rm sys}, \mathcal{P}_{\rm user}),
\end{equation}
where $\mathcal{P}_{\rm sys} = \{\texttt{Role}_i, \texttt{State}_i\}$ indicates the system prompt, encapsulating the agent's role and internal state, while $\mathcal{P}_{\rm user}$ corresponds to the user prompt, which incorporates task specifications, responses from other agents and externally retrieved knowledge.

\subsection{Multi-Agent Communication Pipeline}
Given the task query $\mathcal{Q}$, the multi-agent system conducts collaborative reasoning according to the collaboration graph $\mathcal{G}$, which specifies the pathways for information propagation among the agents. Unlike the conventional message passing schemes in graph neural networks (GNNs) \cite{kipf2017semisupervised,hamilton2017inductive,veličković2018graph}, agent activations within each round follow an execution order derived from a topological sorting of the communication graph, ensuring that all prerequisite inputs are available before an agent is invoked. This collaborative procedure may be repeated for $K$ rounds to enable progressive refinement. Specifically, at round $k$, the mapping function $\phi(\cdot)$ determines the execution index of each agent:
\begin{equation}
    \begin{aligned}
        \phi: \mathcal{G} \mapsto & \sigma, \sigma = [v_{\sigma_1}, v_{\sigma_2}, \cdots, v_{\sigma_N}], \\ 
        & {\rm s.t.} \ \forall i > j, \ v_{\sigma_i} \notin \mathcal{N}_{i}(v_{\sigma_j}), 
    \end{aligned}
\end{equation}
where $\sigma$ specifies the agent execution sequence. $\mathcal{N}_{i}(v_{\sigma_j})$ denotes the in-neighborhood of agent $v_{\sigma_j}$. The execution order guarantees that an agent $v_{\sigma_i}$ is activated only after all agents from which it receives information have complete their execution. Once the execution order is established, each agent generates its response as follows:
\begin{equation}
    \mathcal{P}_i^k = \{\mathcal{P}_{\rm sys}^k, \{\mathcal{Q}, \cup_{v_j \in \mathcal{N}_{i}(v_i)}\mathcal{R}_j^k\}\}, \ \mathcal{R}_i^k = v_i(\mathcal{P}_i^k),
\end{equation}
where the response $\mathcal{R}_i^k$ is conditioned on the system prompt $\mathcal{P}_{\rm sys}^t$ together with a context prompt composed of the query $\mathcal{Q}$ and the incoming messages from neighboring agents. Upon completion of $K$ round communications, an aggregation function is employed to consolidate these responses into the final solution as follows:
\begin{equation}
    \mathcal{S}^K \leftarrow \texttt{Aggregate}(\mathcal{R}_1^K,\mathcal{R}_2^K,\cdots,\mathcal{R}_N^K).
\end{equation}
The \texttt{Aggregate} function can be implemented flexibly and might take various forms, such as majority voting \cite{chen2024more,zhuge2024gptswarm}, concatenating and consolidating responses from all agents for final decision making \cite{zhang2025cut,jiang2023llm}, or directly adopting the output of the last agent $\mathcal{R}_{\sigma_N}^K$ \cite{qian2025scaling,zhang2025gdesigner,li2025assemble}. The communication process may proceed for a fixed number of rounds $K$ or terminate adaptively based on an early-stopping criterion.

\subsection{Automatic Multi-Agent Topology Design}
For a given query $\mathcal{Q}$, a large language model based multi-agent system $\mathcal{G} = (\mathcal{V},\mathcal{E})$ is automatically constructed to satisfy the following design objectives: {\it \textbf{1. Effectiveness}}: maximize the task utility, $u(\mathcal{G}(\mathcal{Q}))$; {\it \textbf{2. Cost Efficiency}}: minimize the financial cost, $c(\mathcal{G;\mathcal{Q}})$; {\it \textbf{3. Adaptiveness}}: adjust the topology in response to varying tasks. The system's objectives can be simultaneously represented in the following optimization framework:
\begin{equation}
   \min_{\mathcal{G} \in \mathbb{G}} \mathcal{L}(\mathcal{G};\mathcal{Q}) = -u(\mathcal{G}(\mathcal{Q})) + \alpha \cdot c(\mathcal{G};\mathcal{Q}),
   \label{eq:utility}
\end{equation}
where function $u(\cdot)$ quantifies task-specific utility (e.g., accuracy, pass@1, etc.), $c(\cdot)$ measures the communication cost, and $\alpha$ is a trade-off hyper-parameter.

\begin{figure*}
    \centering
    \includegraphics[width=0.95\linewidth]{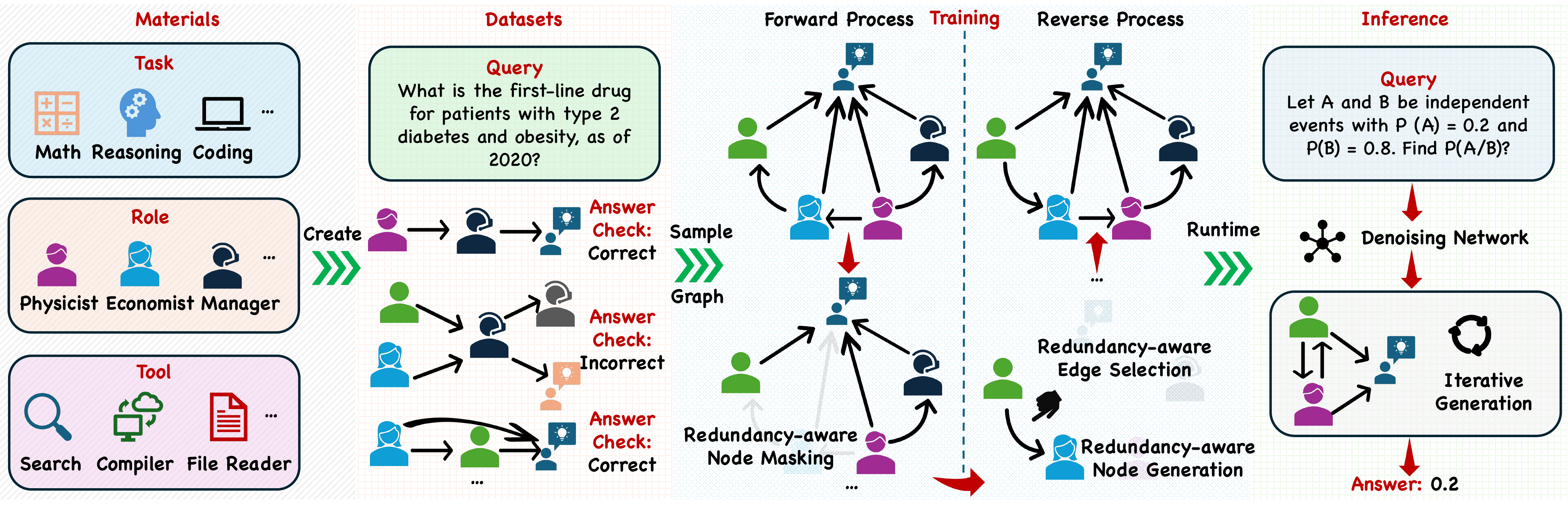}
    \caption{Overview of the RADAR framework. It starts from task-specific inputs and constructs baseline topologies to train a graph diffusion model. The trained denoising network then iteratively refines an initially empty graph to synthesize a task-adaptive topology.}
    \label{fig:model}
    \vspace{-0.2in}
\end{figure*}

\section{The Proposed RADAR Model}
An overview of the proposed RADAR framework is illustrated in Figure \ref{fig:model}, which consists of four principal stages. The workflow starts with task-relevant inputs, including the task query, the set of candidate agents, and available tools, etc. Based on these inputs, multiple baseline topologies are constructed to form a foundational dataset that captures the relationship between communication structures and task performance. This dataset is subsequently used to train graph diffusion models capable of generating high-performing graph topologies. Given a new task, the denoising network progressively refines an initially empty graph through iterative denoising, ultimately synthesizing a task-adaptive communication topology.

\subsection{Effective Size as a Measure of Redundancy}
Effective size is a classical graph theory concept that quantifies structural redundancy within a node's local neighborhood \cite{ronald1992structural}. It characterizes the extent to which a node's connections provide non-overlapping access, discounting neighbors that are highly interconnected and therefore likely lead to redundant information. Inspired by this concept, we adapt effective size to the setting of graph-based multi-agent collaboration, where nodes represent agents and directed edges encode information flow. In this context, we define the effective size of a node's incoming ego network as follows:
\begin{equation}
    \varphi^{i}(v_k) = |\mathcal{N}_{i}(v_k)| - \frac{\sum_{j,q \in {\mathcal{N}_{i}(v_k)}}A_{jq}\mathbb{I}[r(j) = r(q)]}{|\mathcal{N}_{i}(v_k)|},
\end{equation} 
where $\mathcal{N}_i(v_k)$ represents the in-neighborhood of agent $v_k$, while $A_{jq} \in \{0,1\}$ encodes the information flow relationship between agents $v_j$ and $v_q$. The role function $r(\cdot)$ specifies the functional role of each agent, and $\mathbb{I}[\cdot]$ is an indicator function. This equation measures how many distinct and complementary information sources an agent effectively receives. A high incoming effective size indicates that agent $v_k$ benefits from diverse perspectives and complementary expertise, {\it i.e.}, low redundancy. Similarly, the outgoing effective size of agent $v_k$, i.e., $\varphi^o(v_k)$, measures redundancy in information dissemination. An agent is structurally effective when it distributes information across diverse, non-overlapping execution paths, while maintaining sufficient information diversity. Together with the incoming effective size, these two metrics provide a principled characterization of local information efficiency in directed multi-agent collaboration graphs. Formally, we define the effective size for agent $v_k$ in directed communication graphs as follows:
\begin{equation}
    \varphi(v_k) = \varphi^{i}(v_k) \cdot (1 - \beta) + \varphi^{o}(v_k) \cdot \beta,
\end{equation}
where $\beta \in [0,1]$ is a trade-off hyper-parameter that balances the contributions of incoming and outgoing effective size. The value of $\varphi(v_k)$ will play an important role in the communication graph generation process, guiding the model in reasoning over structure information and controlling redundancy.

\subsection{Redundancy-Aware Graph Diffusion}
We formulate the multi-agent communication topology as a directed graph $\mathcal{G} = (\mathcal{V},\mathcal{E})$ characterized by a binary adjacency matrix $\mathbf{A} \in \{0,1\}^{N \times N}$, where $A_{ij} = 1$ indicates the presence of an edge $e_{ij} \in \mathcal{E}$ and $A_{ij} =0$ otherwise. Each edge $e_{ij}$ encodes the directed information flow from agent $v_i$ to agent $v_j$. Under this formulation, the design of multi-agent communication structure is formulated as a graph generation problem, and our goal is to learn a redundancy-aware graph generative model from a collection of training communication graphs via diffusion model.

\textbf{The Forward Process.}
Motivated by the recent advances in discrete graph diffusion models \cite{kong2023autoregressive,chen2023efficient,yang2023directional}, we introduce a forward diffusion process that progressively masks nodes together with their connected edges. To guide this process, we design a redundancy-aware ordering network that prioritizes nodes based on their effective size, thereby imposing structured regularities that simplify generative learning. Intuitively, graphs with high effective size tend to decompose into weakly overlapping substructures, making it more tractable to be generated incrementally. Accordingly, we employ an ordering network $q_{\psi}(\pi|\mathcal{G}_0,\varphi)$ that samples a node $v_{\pi(t)}$ to be masked at each diffusion step $t$, yielding the corresponding partially masked graph $\mathcal{G}_t$. Under this formulation, the ordering network operates in a recurrent manner across diffusion steps as follows:
\begin{equation}
    q_{\psi}(\pi|\mathcal{G}_0,\varphi) = \prod_t q_{\psi}(\pi_t|\mathcal{G}_0, \varphi, \pi_{(< t)}).
\end{equation}
Specifically, at diffusion step $t$, the selection of the $t$-th node $\pi_t$ is modeled as a conditional distribution that depends on the original graph $\mathcal{G}_0$, the node-level effective size $\varphi(v_t)$, and the previously sampled ordering $\pi_{(< t)}$. The graph structure is subsequently processed by a graph neural network (GNN) \cite{kipf2017semisupervised,hamilton2017inductive,zhang2021h2mn} to obtain node-level representations. To explicitly capture the partial orderings, positional encodings \cite{vaswani2017attention} are incorporated into the node features prior to message passing. Let ${h}_t$ denote the resulting embedding of node $v_t$ produced by GNN. The conditional distribution $q_{\psi}(\pi_t|\mathcal{G}_0, \varphi, \pi_{(< t)})$ is then parameterized as a categorical distribution over nodes:
\begin{equation}
    q_{\psi}(\pi_t|\mathcal{G}_0, \varphi, \pi_{(< t)}) = \frac{{\rm exp}({h}_t + \varphi(v_t))}{\sum_{j \notin \pi_{(<t)}}{\rm exp}({h}_j)},
\end{equation}
where the output is a scalar score for candidate node $v_t$, representing the probability of being selected at step $t$. Through this formulation, the ordering network could sequentially select nodes in a structure-aware and redundancy-informed manner.

\textbf{The Reverse Process.}
During the reverse generative phase, a denoising network $p_{\theta}(\mathcal{G}_{t}|\mathcal{G}_{t+1},\mathcal{Q})$ progressively reconstructs the graph by inverting the forward diffusion process conditioning on the query $\mathcal{Q}$, thereby enabling the generated topology to be task-adaptive. At step $t$, the denoising network takes the partially masked graph $\mathcal{G}_{t+1}$ as input and maps each node $v_i$ into a latent embedding space. The node representations are then iteratively refined through graph neural network message-passing operations. Specifically, at the $(l+1)$-th message passing layer, the embedding of node $v_i$ is updated through aggregating attention-weighted messages from its incoming neighbors:
\begin{equation}
    \alpha_{i,j} = \frac{{\rm exp}({\rm ReLU}(\mathbf{a}^{\top}[\mathbf{Wh}_i^{l}||\mathbf{Wh}_j^{l}]))}{\sum_{k \in \mathcal{N}_i}{\rm exp}({\rm ReLU}(\mathbf{a}^{\top}[\mathbf{Wh}_i^{l}||\mathbf{Wh}_k^{l}]))},
\end{equation}
\begin{equation}
    \mathbf{h}_i^{l+1} = {\rm ReLU}(\sum_{j \in \mathcal{N}_i}\alpha_{i,j}\mathbf{Wh}_j^{l}).
\end{equation}
Here, $\mathbf{W}$ denotes the learnable weight matrix and $\mathbf{a}$ represents the attention vector. $\mathcal{N}_i$ is the neighborhood of node $v_i$, while ReLU refers to the activation function. After passing message $L$ rounds, the denoising network produces the final embedding $\mathbf{h}^L_i$ for each node. We then apply a bias term based on the effective size, updating the representation as $\mathbf{h}_i^L = \mathbf{h}_i^L + \varphi(v_i)\mathbf{1}$. Based on these embeddings, multi-layer perceptions are utilized to predict the agent role of the newly recovered node $v_{\pi_t}$ and its connectivity to the set of previously denoised nodes $\{v_{\pi (> t)}\}$. Rather than generating edges in an autoregressive manner, the connections between $v_{\pi_t}$ and all existing nodes are inferred jointly using a mixture of multinomial distributions. This design captures dependencies among edge variables while reducing the number of generation steps to $\mathcal{O}(N)$, which enables efficient graph structure generation.

\subsection{Training Objective}
To optimize the model's parameters, we use a reinforcement learning based training strategy to jointly update the diffusion ordering network $q_{\psi}(\pi|\mathcal{G}_0,\varphi)$ and the denoising network $p_{\theta}(\mathcal{G}_t|\mathcal{G}_{t+1},\mathcal{Q})$ using gradient descent. At each training iteration, for the $i$-th training graph $\mathcal{G}_0^{(i)}$, we generate $M$ diffusion trajectories by sampling node-masking orderings $\pi^{i,m}$ from the ordering network. Each trajectory consists of a sequence of partially masked graphs $\{\mathcal{G}_t^{i,m}\}_{ 1\le t \le N}$, where $N$ denotes the number of nodes. For each trajectory, diffusion steps are executed over $T$ time steps. Conditioned on these sampled trajectories, the denoising network is trained to minimize the negative variational lower bound using stochastic gradient descent. For notation clarity, the superscript $i$ is omitted in the following formulation:
\begin{equation}
    \nabla_{\theta}\mathcal{G} = \sum_{m,t}\sum_{k \in \pi(\le t)}w_{k}^{m} \nabla {\rm log} \ p_{\theta}(\mathcal{G}_{v_k}^{\pi(>t)}|\mathcal{G}_{t+1}^{m},\mathcal{Q}),
\end{equation}
where $w_k^{m} = q_{\psi}(\pi_t^{m} = k|\mathcal{G}_0,\varphi,\pi_{(<t)}^{m})$ denotes the probability assigned by ordering network to selecting node $v_k$ at diffusion step $t$. The term $p_{\theta}(\mathcal{G}_{v_k}^{\pi(>t)}|\mathcal{G}_{t+1}^{m},\mathcal{Q})$ represents the denoising network's conditional distribution for jointly generating node $v_k$ and its connections to previously reconstructed nodes. The resulting weighted log-likelihood gradient specifies how the denoising model updates its parameters to favor the generation of structurally coherent nodes and edges while adhering to the reverse order of the diffusion process.

For the diffusion ordering network, it is optimized using a standard reinforcement learning technique, i.e., the REINFORCE algorithm \cite{williams1992simple}, since the node masking orderings it produces are discrete and non-differentiable:
\begin{equation}
    \nabla_{\psi}\mathcal{G} = \sum_{m}R^{m}\nabla{\rm log} \ q_{\psi}(\pi|\mathcal{G}_0^{m},\varphi),
\end{equation}
where $R^m$ provides a learning signal for the ordering network by measuring how well a sampled node-masking sequence supports high-likelihood reconstruction in the denoising model:
\begin{equation}
    R^{m}=-\sum_t\sum_{k \in \pi(\le t)}w_k^{m}{\rm log} \ p_{\theta}(\mathcal{G}_{v_k}^{\pi(>t)}|\mathcal{G}_{t+1}^{m},\mathcal{Q}).
\end{equation}
In addition to structural fidelity, we explicitly optimize for downstream task utility. However, the utility function $u(\cdot)$ is often non-differentiable and computationally intractable, as it often relies on external API evaluations or black-box task executions \cite{li2023api}. To this end, we adopt a policy gradient based estimator to optimize Equation (\ref{eq:utility}) as follows:
\begin{equation}
    \nabla_{\theta}\mathbb{E}[\mathcal{G}] \approx \frac{1}{\mathcal{B}}\sum_{k=1}^{\mathcal{B}}u(\mathcal{G}^{(k)}(\mathcal{Q}))\nabla_{\theta}{\rm log} \ p_{\theta}(\mathcal{G}^{(k)}|\mathcal{Q}).
    \label{eq:utility_update}
\end{equation}
This formulation enables the model to directly leverage task-level feedback to guide topology generation, thereby encouraging agent communication structures that are aligned with task performance objectives. In practice, we adopt a periodic and subsampled utility evaluation strategy. The task utility is computed only for a subset of generated graphs at a fixed update frequency across training epochs and mini-batches. Through the joint updating of its key components, the proposed model enables the fully automated generation of multi-agent collaboration structures.

\section{Experiments}

\subsection{Datasets}
To comprehensively evaluate our proposed model, we conduct experiments across a diverse set of datasets including MMLU \cite{hendrycks2021measuring} ({\it general reasoning}), GSM8K \cite{cobbe2021training}, MultiArith \cite{roy2015solving}, SVAMP \cite{patel2021nlp}, AQuA \cite{ling2017program} ({\it mathematical problem solving}), and HumanEval \cite{chen2021evaluatinglargelanguagemodels} ({\it code generation}). The dataset statistics and evaluation metrics are presented in Table \ref{tab:datasets} in the Appendix \ref{ap:dataset}.

\subsection{Baselines}
We compare RADAR against three types of baselines: (1) \textbf{Single Agent Methods} including CoT \cite{wei2022chain}, ComplexCoT \cite{fu2023complexitybased}, SC \cite{wang2023selfconsistency}; (2) \textbf{Multi-Agent Systems} including MultiPersona \cite{wang2024unleashing}, LLM-Debate \cite{du2023improving}, LLM-Blender \cite{jiang2023llm}, DyLAN \cite{liu2024dynamic}, AgentVerse \cite{chen2024agentverse}, MacNet \cite{qian2025scaling}; (3) \textbf{Autonomous Multi-Agent Systems} including AutoAgents \cite{chen2024autoagents}, GPTSwarm \cite{zhuge2024gptswarm}, ADAS \cite{hu2025automated}, AgentSquare \cite{shang2025agentsquare}, AFlow \cite{zhang2025aflow}, G-Designer \cite{zhang2025gdesigner}, AgentPrune \cite{zhang2025cut}, GTD \cite{jiang2025dynamic}, MaAS \cite{zhang2025multiagent}, ARG-Designer \cite{li2025assemble}. Additional details are provided in Appendix \ref{ap:baselines}.

\subsection{Implementation Details}
We interact with GPT models through the OpenAI API\footnote{https://api.openai.com/v1/chat/completions}, with experiments primarily conducted on \texttt{gpt-4o-mini}. For all runs, the temperature is fixed at $0.2$, and the maximum token budget per execution is set to $1,000$. Query representations are encoded using the \texttt{all-MiniLM-L6-v2} model \cite{wang2020minilm}, which outputs embeddings of dimension $384$. Across all datasets, we sample $50$ queries for model training. The initial diffusion dataset is constructed by instantiating a diverse set of baseline communication topologies (e.g., \texttt{fully connected}, \texttt{mesh}, \texttt{star}, \texttt{layered}, and \texttt{random}) with varying numbers of agents (e.g., 3 or 4 agents). For each configuration, we randomly assign agent roles and evaluate the resulting multi-agent system on sampled training instances (e.g., whether the final answer is correct or incorrect). This provides a diverse set of graph-structured samples to initialize the diffusion model. The ordering network is implemented as a three-layer RGCN \cite{schlichtkrull2018modeling}, while the denoising network adopts a three-layer GAT \cite{veličković2018graph} with hidden dimension setting as $32$. The parameters are optimized using Adam \cite{kingma2014adam} with learning rates of $5 \times 10^{-4}$ for the diffusion-based ordering network and $1 \times 10^{-4}$ for the denoising network. The hyperparameters $\alpha$ and $\beta$ are tuned via grid search over the interval $[0, 1]$.

\subsection{Performance Analysis}
{
\rowcolors{2}{gray!10}{}
\begin{table*}
  \centering
   \caption{Performance comparison across three categories of baselines, including single-agent methods, multi-agent systems, and autonomous multi-agent systems. The best results are highlighted in bold. All methods, except those in the single-agent category, employ \textbf{five} \texttt{gpt-4o-mini}-based agents. The ``Auto" column indicates whether a method supports autonomous topology design.}
  \begin{tabular}{lcccccccc}
    \Xhline{1.0pt}
    \rowcolor{blue!12}
    \textbf{Methods} & \textbf{Auto} & \textbf{MMLU} & \textbf{GSM8K} & \textbf{MultiArith} & \textbf{SVAMP} & \textbf{AQuA} & \textbf{HumanEval} & \textbf{Avg.} \\
    \Xhline{1.0pt}
    Vanilla & \textcolor{red!70!black}{\ding{55}} & 78.54 & 87.45 & 96.85 & 86.67 & 78.92 & 87.08  & 85.92  \\
    \Xhline{0.6pt}
    CoT & \textcolor{red!70!black}{\ding{55}} & 79.26$_{\textcolor{orange}{\uparrow 0.72}}$ & 87.10$_{\textcolor{blue}{\downarrow 0.35}}$ & 96.31$_{\textcolor{blue}{\downarrow 0.54}}$ & 87.33$_{\textcolor{orange}{\uparrow 0.66}}$ & 75.20$_{\textcolor{blue}{\downarrow 3.72}}$ & 88.13$_{\textcolor{orange}{\uparrow 1.05}}$ & 85.55 \\
    ComplexCoT & \textcolor{red!70!black}{\ding{55}} & 79.80$_{\textcolor{orange}{\uparrow 1.26}}$  & 86.89$_{\textcolor{blue}{\downarrow 0.56}}$ & 96.70$_{\textcolor{blue}{\downarrow 0.15}}$ & 87.67$_{\textcolor{orange}{\uparrow 1.00}}$ & 75.59$_{\textcolor{blue}{\downarrow 3.33}}$ & 87.49$_{\textcolor{orange}{\uparrow 0.41}}$ & 85.69 \\
    SC (COT$\times$5) & \textcolor{red!70!black}{\ding{55}} & 80.66$_{\textcolor{orange}{\uparrow 2.12}}$ & 87.57$_{\textcolor{orange}{\uparrow 0.12}}$ & 96.58$_{\textcolor{blue}{\downarrow 0.27}}$ & 88.00$_{\textcolor{orange}{\uparrow 1.33}}$ & 82.28$_{\textcolor{orange}{\uparrow 3.36}}$ & 88.60$_{\textcolor{orange}{\uparrow 1.52}}$ & 87.28 \\
    \Xhline{0.6pt}
    MultiPersona & \textcolor{red!70!black}{\ding{55}} & 77.69$_{\textcolor{blue}{\downarrow 0.85}}$ & 87.50$_{\textcolor{orange}{\uparrow 0.05}}$ & 97.49$_{\textcolor{orange}{\uparrow 0.64}}$ & 87.00$_{\textcolor{orange}{\uparrow 0.33}}$ & 79.23$_{\textcolor{orange}{\uparrow 0.31}}$ & 88.32$_{\textcolor{orange}{\uparrow 1.24}}$ & 86.21 \\
    LLM-Debate & \textcolor{red!70!black}{\ding{55}} & 80.56$_{\textcolor{orange}{\uparrow 2.02}}$ & 89.47$_{\textcolor{orange}{\uparrow 2.02}}$ & 97.33$_{\textcolor{orange}{\uparrow 0.48}}$ & 89.00$_{\textcolor{orange}{\uparrow 2.33}}$ & 79.70$_{\textcolor{orange}{\uparrow 0.78}}$ & 88.68$_{\textcolor{orange}{\uparrow 1.60}}$ & 87.46 \\
    LLM-Blender & \textcolor{red!70!black}{\ding{55}} & 80.29$_{\textcolor{orange}{\uparrow 1.75}}$  & 88.35$_{\textcolor{orange}{\uparrow 0.90}}$ & 97.29$_{\textcolor{orange}{\uparrow 0.44}}$ & 87.33$_{\textcolor{orange}{\uparrow 0.66}}$ & 78.99$_{\textcolor{orange}{\uparrow 0.07}}$ & 88.80$_{\textcolor{orange}{\uparrow 1.72}}$ & 86.84\\
    DyLAN & \textcolor{red!70!black}{\ding{55}} & 79.86$_{\textcolor{orange}{\uparrow 1.32}}$ & 89.98$_{\textcolor{orange}{\uparrow 2.53}}$ & 97.12$_{\textcolor{orange}{\uparrow 0.27}}$ & 88.67$_{\textcolor{orange}{\uparrow 2.00}}$ & 79.59$_{\textcolor{orange}{\uparrow 0.67}}$ & 90.42$_{\textcolor{orange}{\uparrow 3.34}}$ & 87.61 \\
    AgentVerse & \textcolor{red!70!black}{\ding{55}} & 78.39$_{\textcolor{blue}{\downarrow 0.15}}$ & 89.91$_{\textcolor{orange}{\uparrow 2.46}}$ & 97.50$_{\textcolor{orange}{\uparrow 0.65}}$ & 88.33$_{\textcolor{orange}{\uparrow 1.66}}$ & 77.47$_{\textcolor{blue}{\downarrow 1.45}}$ & 89.29$_{\textcolor{orange}{\uparrow 2.21}}$ & 86.82 \\
    MacNet & \textcolor{red!70!black}{\ding{55}} & 79.55$_{\textcolor{orange}{\uparrow 1.01}}$  & 87.95$_{\textcolor{orange}{\uparrow 0.50}}$ & 96.03$_{\textcolor{blue}{\downarrow 0.82}}$ & 86.00$_{\textcolor{blue}{\downarrow 0.67}}$ & 79.23$_{\textcolor{orange}{\uparrow 0.31}}$ & 84.57$_{\textcolor{blue}{\downarrow 2.51}}$ & 85.55\\
    \Xhline{0.6pt}
    AutoAgents & \textcolor{teal}{\ding{51}} & 79.59$_{\textcolor{orange}{\uparrow 1.05}}$ & 87.69$_{\textcolor{orange}{\uparrow 0.24}}$ & 96.42$_{\textcolor{blue}{\downarrow 0.43}}$ & 86.34$_{\textcolor{blue}{\downarrow 0.33}}$ & 78.65$_{\textcolor{blue}{\downarrow 0.27}}$ & 87.64$_{\textcolor{orange}{\uparrow 0.56}}$ & 86.05 \\
    GPTSwarm & \textcolor{teal}{\ding{51}} & 78.36$_{\textcolor{blue}{\downarrow 0.18}}$ & 89.14$_{\textcolor{orange}{\uparrow 1.69}}$ & 96.79$_{\textcolor{blue}{\downarrow 0.06}}$ & 88.67$_{\textcolor{orange}{\uparrow 2.00}}$ & 80.71$_{\textcolor{orange}{\uparrow 1.79}}$ & 89.32$_{\textcolor{orange}{\uparrow 2.24}}$ & 87.17\\
    ADAS & \textcolor{teal}{\ding{51}} & 78.39$_{\textcolor{blue}{\downarrow 0.15}}$ & 86.12$_{\textcolor{blue}{\downarrow 1.33}}$ & 96.02$_{\textcolor{blue}{\downarrow 0.83}}$ & 86.33$_{\textcolor{blue}{\downarrow 0.34}}$ & 77.71$_{\textcolor{blue}{\downarrow 1.21}}$ & 84.19$_{\textcolor{blue}{\downarrow 2.89}}$ & 84.79\\
    AgentSquare &  \textcolor{teal}{\ding{51}} & 79.58$_{\textcolor{orange}{\uparrow 1.04}}$ & 87.62$_{\textcolor{orange}{\uparrow 0.17}}$ & 97.77$_{\textcolor{orange}{\uparrow 0.92}}$ & 88.00$_{\textcolor{orange}{\uparrow 1.33}}$ & 81.50$_{\textcolor{orange}{\uparrow 2.58}}$ & 89.08$_{\textcolor{orange}{\uparrow 2.00}}$ & 87.26 \\
    AFlow &  \textcolor{teal}{\ding{51}} & 81.80$_{\textcolor{orange}{\uparrow 3.26}}$ & 91.16$_{\textcolor{orange}{\uparrow 3.71}}$ & 96.22$_{\textcolor{blue}{\downarrow 0.63}}$ & 88.33$_{\textcolor{orange}{\uparrow 1.66}}$ & 80.90$_{\textcolor{orange}{\uparrow 1.98}}$ & 90.93$_{\textcolor{orange}{\uparrow 3.85}}$ & 88.22 \\
    G-Designer &  \textcolor{teal}{\ding{51}} & 80.39$_{\textcolor{orange}{\uparrow 1.85}}$ & 91.09$_{\textcolor{orange}{\uparrow 3.64}}$ & 97.78$_{\textcolor{orange}{\uparrow 0.93}}$ & 90.00$_{\textcolor{orange}{\uparrow 3.33}}$ & 80.75$_{\textcolor{orange}{\uparrow 1.83}}$ & 89.37$_{\textcolor{orange}{\uparrow 2.29}}$ & 88.23 \\
    AgentPrune &  \textcolor{teal}{\ding{51}} & 82.40$_{\textcolor{orange}{\uparrow 3.86}}$ & 91.92$_{\textcolor{orange}{\uparrow 4.47}}$ & 97.88$_{\textcolor{orange}{\uparrow 1.03}}$ & 90.37$_{\textcolor{orange}{\uparrow 3.70}}$ & 80.93$_{\textcolor{orange}{\uparrow 2.01}}$ & 87.17$_{\textcolor{orange}{\uparrow 0.09}}$ & 88.22\\
    GTD & \textcolor{teal}{\ding{51}} & 79.41$_{\textcolor{orange}{\uparrow 0.87}}$ & 91.38$_{\textcolor{orange}{\uparrow 3.93}}$ & 96.20$_{\textcolor{blue}{\downarrow 0.65}}$ & 90.24$_{\textcolor{orange}{\uparrow 3.57}}$ & 79.44$_{\textcolor{orange}{\uparrow 0.52}}$ & 87.86$_{\textcolor{orange}{\uparrow 0.78}}$ & 87.42\\
    MaAS & \textcolor{teal}{\ding{51}} & 82.32$_{\textcolor{orange}{\uparrow 3.78}}$ & 91.13$_{\textcolor{orange}{\uparrow 3.68}}$ & 98.08$_{\textcolor{orange}{\uparrow 1.23}}$ & 89.65$_{\textcolor{orange}{\uparrow 2.98}}$ & 80.25$_{\textcolor{orange}{\uparrow 1.33}}$ & 89.57$_{\textcolor{orange}{\uparrow 2.49}}$ & 88.50 \\
    ARG-Designer & \textcolor{teal}{\ding{51}} & 79.10$_{\textcolor{orange}{\uparrow 0.56}}$ & 91.25$_{\textcolor{orange}{\uparrow 3.80}}$ & 98.55$_{\textcolor{orange}{\uparrow 1.70}}$ & 92.21$_{\textcolor{orange}{\uparrow 5.54}}$ & 81.10$_{\textcolor{orange}{\uparrow 2.18}}$ & 89.19$_{\textcolor{orange}{\uparrow 2.11}}$ & 88.57\\
    \Xhline{0.6pt}
    RADAR & \textcolor{teal}{\ding{51}} & \textbf{83.66$_{\textcolor{orange}{\uparrow 5.12}}$}  & \textbf{92.51$_{\textcolor{orange}{\uparrow 5.06}}$} & \textbf{98.81$_{\textcolor{orange}{\uparrow 1.96}}$} & \textbf{93.26$_{\textcolor{orange}{\uparrow 6.59}}$} & \textbf{82.84$_{\textcolor{orange}{\uparrow 3.92}}$} & \textbf{91.28$_{\textcolor{orange}{\uparrow 4.20}}$} & \textbf{90.32} \\
    \Xhline{1.0pt}
  \end{tabular}
  \vspace{-0.1in}
  \label{tab:mainresutls}
\end{table*}
}

The experimental results reported in Table \ref{tab:mainresutls} demonstrate that our proposed RADAR is highly effective at designing multi-agent collaboration topologies. Specifically, RADAR consistently achieves the strongest performance across the six benchmarks. In particular, the multi-agent systems generated by RADAR outperform the vanilla single-agent baseline by margins ranging from $1.96\%$ to $6.59\%$. When compared with the recent strongest learning-based baselines, such as ARG-Designer, RADAR delivers an average performance improvement of $1.75\%$. Importantly, our results indicate that multi-agent collaboration does not always guarantee superior performance: several existing multi-agent topologies fail to consistently outperform single-agent systems like GPTSwarm and ADAS. This is largely because poorly coordinated agent interactions can introduce redundant reasoning, conflicting intermediate conclusions, or ineffective information aggregation, which might offset the potential benefits of collaboration. On the other hand, some existing multi-agent methods achieve strong empirical performance but suffer from substantial token consumption overhead, as shown in Figure \ref{fig:token}. In contrast, RADAR attains superior accuracy with lower computational cost, illustrating a strong balance between effectiveness and token consumption.

Furthermore, we observe that the effectiveness of different systems varies substantially across evaluation scenarios with different levels of task difficulty. On relatively simpler reasoning benchmarks such as GSM8K, most baselines exhibit significant performance improvements over the single-agent setting, indicating that even straightforward forms of collaboration can be beneficial in low-complexity tasks. However, this performance gap narrows considerably in more challenging scenarios. For instance, in the most difficult code generation tasks, the majority of baseline methods achieve only modest gains of approximately $3\%$, suggesting limited capability of their collaboration strategies under increased task complexity. In contrast, RADAR continues to deliver consistent benefits, achieving a $4.2\%$ improvement in this setting. These results highlight RADAR’s superior ability to construct effective collaboration topologies that remain robust as task difficulty increases, further underscoring its advantage over existing multi-agent approaches.
  
\subsection{Token Economical}

\begin{figure}
    \centering
    \begin{subfigure}{0.49\linewidth}
        \includegraphics[width=\linewidth]{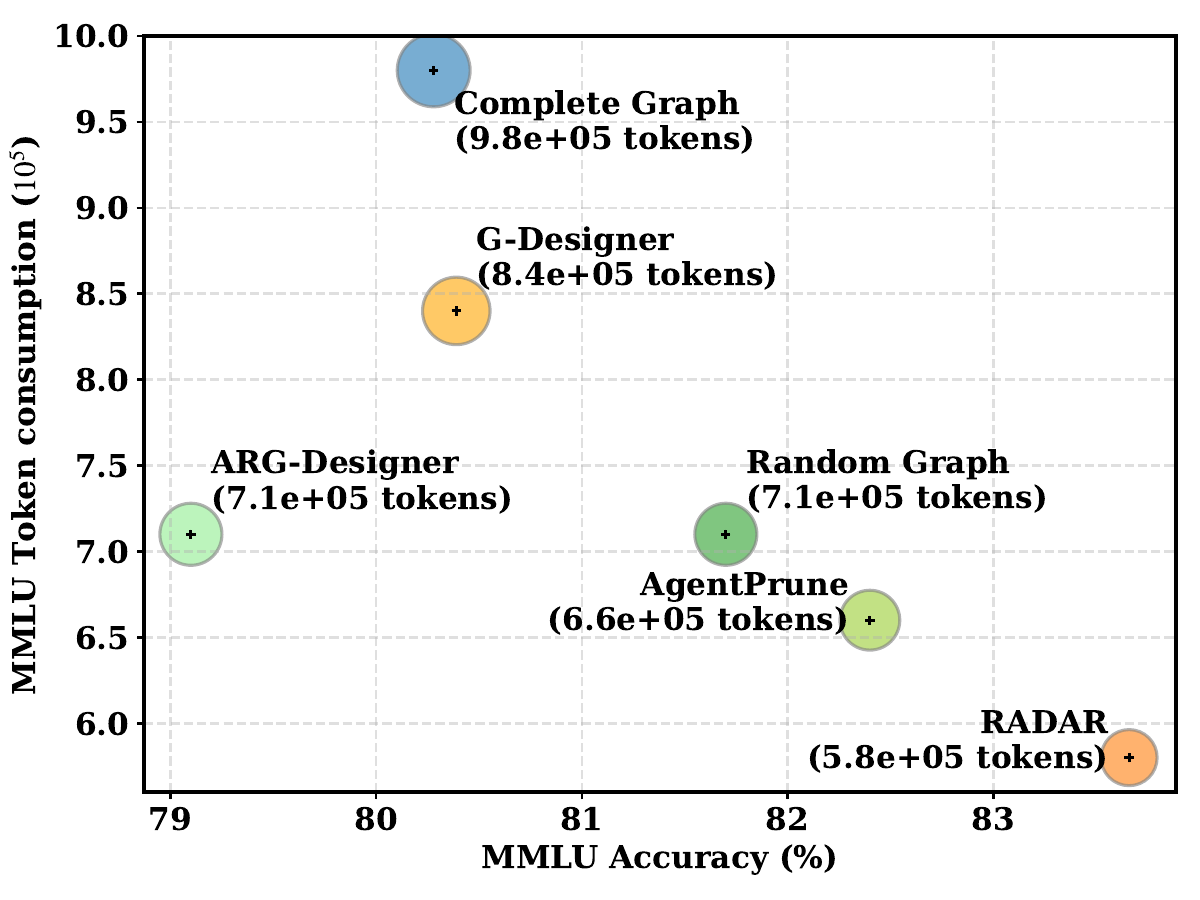} 
    \end{subfigure}
    \hfill
    \begin{subfigure}{0.49\linewidth}
        \includegraphics[width=\linewidth]{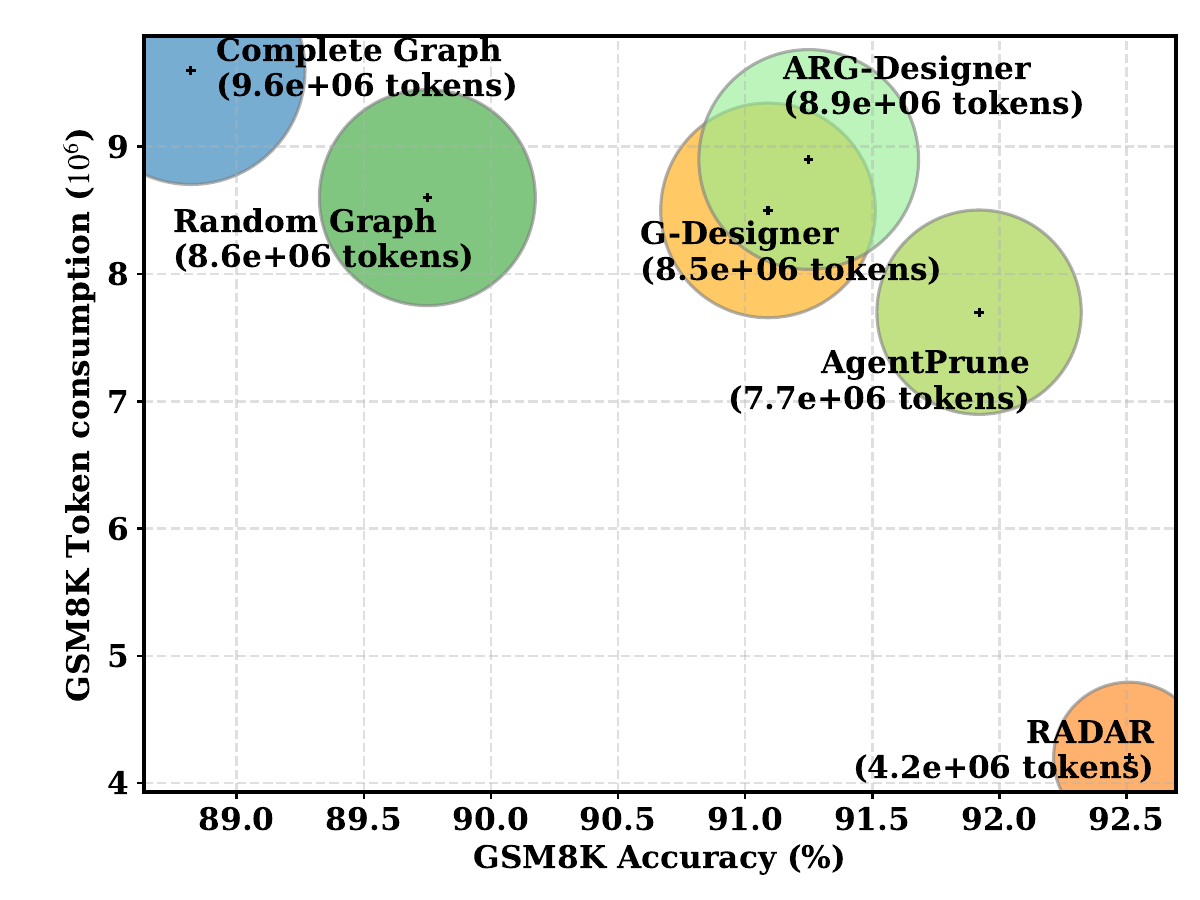} 
    \end{subfigure}
    \caption{Visualization of the performance and token consumption.}
    \label{fig:token}
\end{figure}

One key advantage of RADAR lies in its capability to generate task-specific collaboration topologies, thereby avoiding unnecessary complexity and reducing token consumption. Figure \ref{fig:token} illustrates the trade-off between model performance and token usage. Our proposed RADAR demonstrates a favorable trade-off, maintaining strong performance while exhibiting high token efficiency. On the GSM8K dataset, RADAR is the most token-efficient approach, consuming only $4.2 \times 10^6$ tokens (including both prompt tokens and completion tokens), approximately half the token cost of GDesigner, while attaining the highest performance. This advantage becomes more pronounced as the number of evaluation samples increases, reflecting RADAR’s task-adaptive design. In contrast, more complex communication structures, such as fully connected graphs, incur substantially higher token costs. By explicitly accounting for redundancy and exploring a fine-grained topology, RADAR attains superior token economy compared to existing approaches.

\subsection{Robustness Analysis}

\begin{figure}
    \centering
    \includegraphics[width=\linewidth]{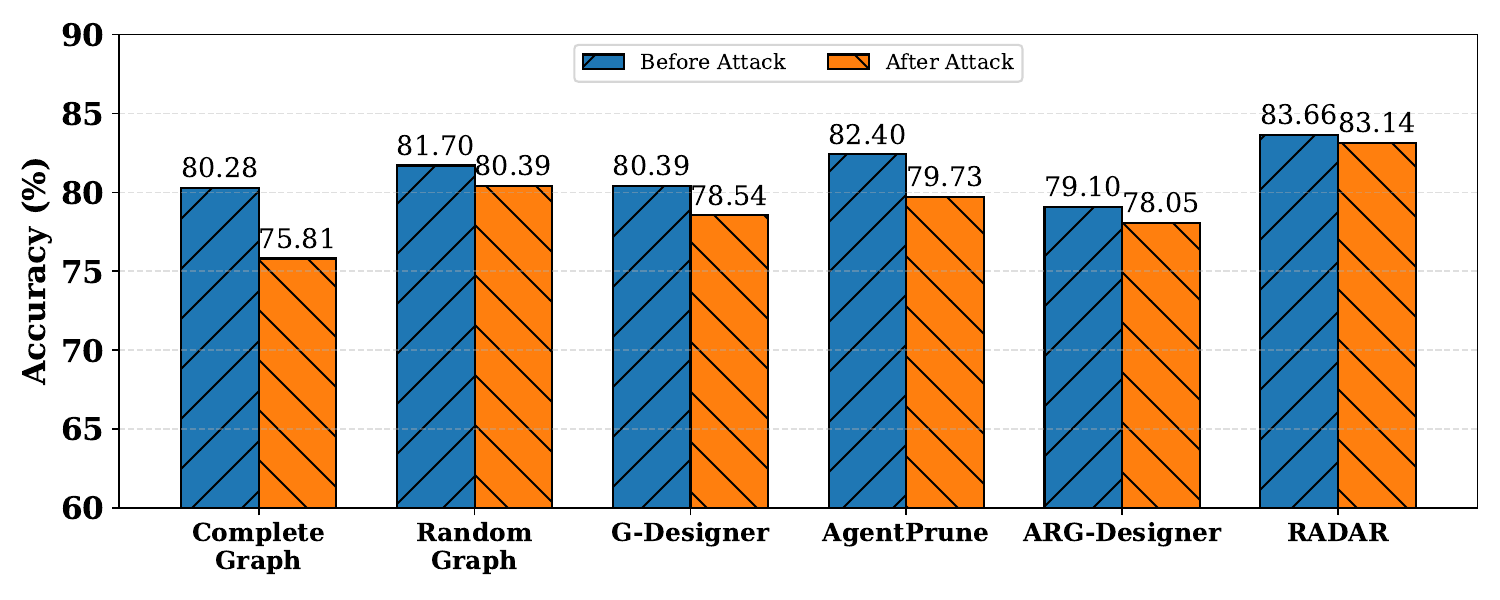}
    \caption{Before and after prompt attacks on MMLU dataset.}
    \label{fig:attack}
    \vspace{-0.2in}
\end{figure}

Following prior work \cite{zhuge2024gptswarm, zhang2025gdesigner}, we inject system prompt attacks into two of the five agents in the collaboration framework. More specifically, we compromise the role prompt of selected agents by altering their roles to a liar agent that intentionally provides false information. As demonstrated in Figure \ref{fig:attack}, while the degree of degradation varies across methods, most frameworks exhibit substantial performance drops under such attacks. For example, the complete graph topology suffers significant degradation, with performance decreasing by up to $4.47\%$, while ARG-Designer demonstrates relatively strong robustness, with only a modest performance drop of approximately $1.05\%$. In contrast, RADAR exhibits exceptional resilience to adversarial perturbations, maintaining nearly identical performance before and after the attack.

\subsection{Ablation Studies}
\textbf{Impact of the key components.} We evaluate three variants of our proposed RADAR across three datasets. Specifically, (1) {\it \textbf{w/o ES}} removes the effective size component from both the ordering network and the denoising network; (2) {\it \textbf{w/o utility}} eliminates the utility loss defined in Equation (\ref{eq:utility}); (3) {\it \textbf{w/o query}} removes the task query input from the denoising network; (4) {\it \textbf{ON w/o ES}} and {\it \textbf{DN w/o ES}} removes the effective size component from the ordering and denoising network respectively and (5) {\it \textbf{non-diffusion}} denotes a baseline that employs non-diffusion techniques (e.g., ARG-Designer). As shown in Table \ref{tab:ablation}, removing any component consistently degrades the model's performance. In particular, excluding the task query impairs the model’s task adaptiveness, while removing the effective size component results in a systematic decline in performance. Similarly, replacing the diffusion-based generation mechanism also leads to noticeable degradation, highlighting its importance in modeling structural dependencies. These results empirically validate the effectiveness of our design. 

\begin{table}
    \centering
    \caption{Ablation studies on three datasets. ``ON" denotes ordering network and ``DN" indicates denoising network.}
    \begin{tabular}{lccc}
        \Xhline{1.0pt}
         \textbf{Variants} & \textbf{MMLU} & \textbf{GSM8K} & \textbf{MultiArith}  \\
        \Xhline{1.0pt}
         RADAR &  83.66 & 92.51 & 98.81 \\
         \Xhline{1.0pt}
         {\it w/o} ES & 81.05$_{\textcolor{blue}{\downarrow 2.61}}$ & 91.22$_{\textcolor{blue}{\downarrow 1.29}}$ & 98.31$_{\textcolor{blue}{\downarrow 0.50}}$ \\
         {\it w/o} utility & 82.96$_{\textcolor{blue}{\downarrow 0.70}}$ & 92.02$_{\textcolor{blue}{\downarrow 0.49}}$ & 98.47$_{\textcolor{blue}{\downarrow 0.34}}$ \\
         {\it w/o} query & 79.08$_{\textcolor{blue}{\downarrow 4.58}}$ & 91.82$_{\textcolor{blue}{\downarrow 0.69}}$ & 97.81$_{\textcolor{blue}{\downarrow 1.00}}$ \\
         ON {\it w/o} ES & 79.74$_{\textcolor{blue}{\downarrow 3.92}}$ & 91.96$_{\textcolor{blue}{\downarrow 0.55}}$ & 98.47$_{\textcolor{blue}{\downarrow 0.34}}$ \\
         DN {\it w/o} ES & 80.39$_{\textcolor{blue}{\downarrow 3.27}}$ & 92.37$_{\textcolor{blue}{\downarrow 0.14}}$ & 98.01$_{\textcolor{blue}{\downarrow 0.80}}$ \\
         non-diffusion & 79.10$_{\textcolor{blue}{\downarrow 4.56}}$ & 91.25$_{\textcolor{blue}{\downarrow 1.26}}$ & 98.55$_{\textcolor{blue}{\downarrow 0.26}}$ \\
        \Xhline{1.0pt}
    \end{tabular}
    \label{tab:ablation}
    \vspace{-0.1in}
\end{table}

\begin{figure}
    \centering
    \begin{subfigure}{0.49\linewidth}
        \includegraphics[width=\linewidth]{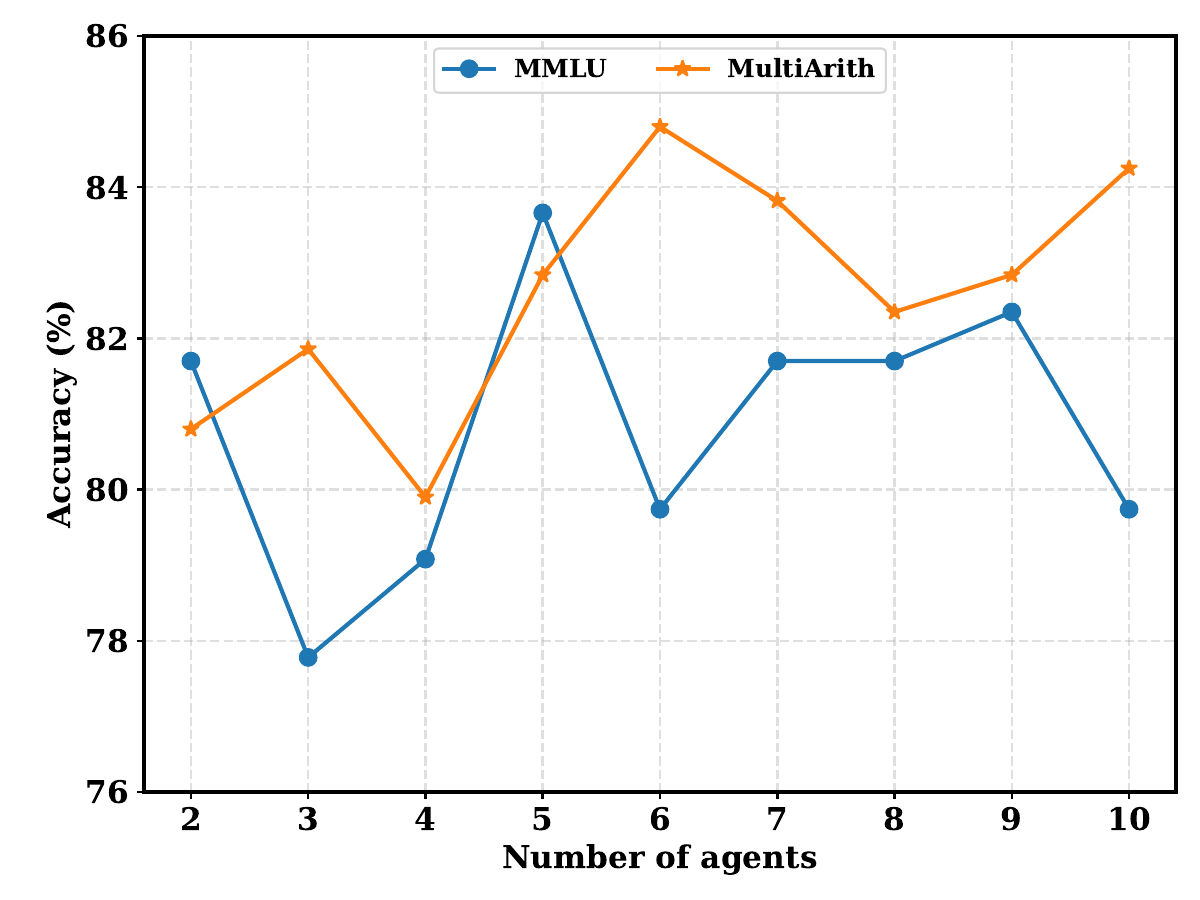} 
        \caption{Impact of agent numbers.}
        \label{fig:agent_num}
    \end{subfigure}
    \hfill
    \begin{subfigure}{0.49\linewidth}
        \includegraphics[width=\linewidth]{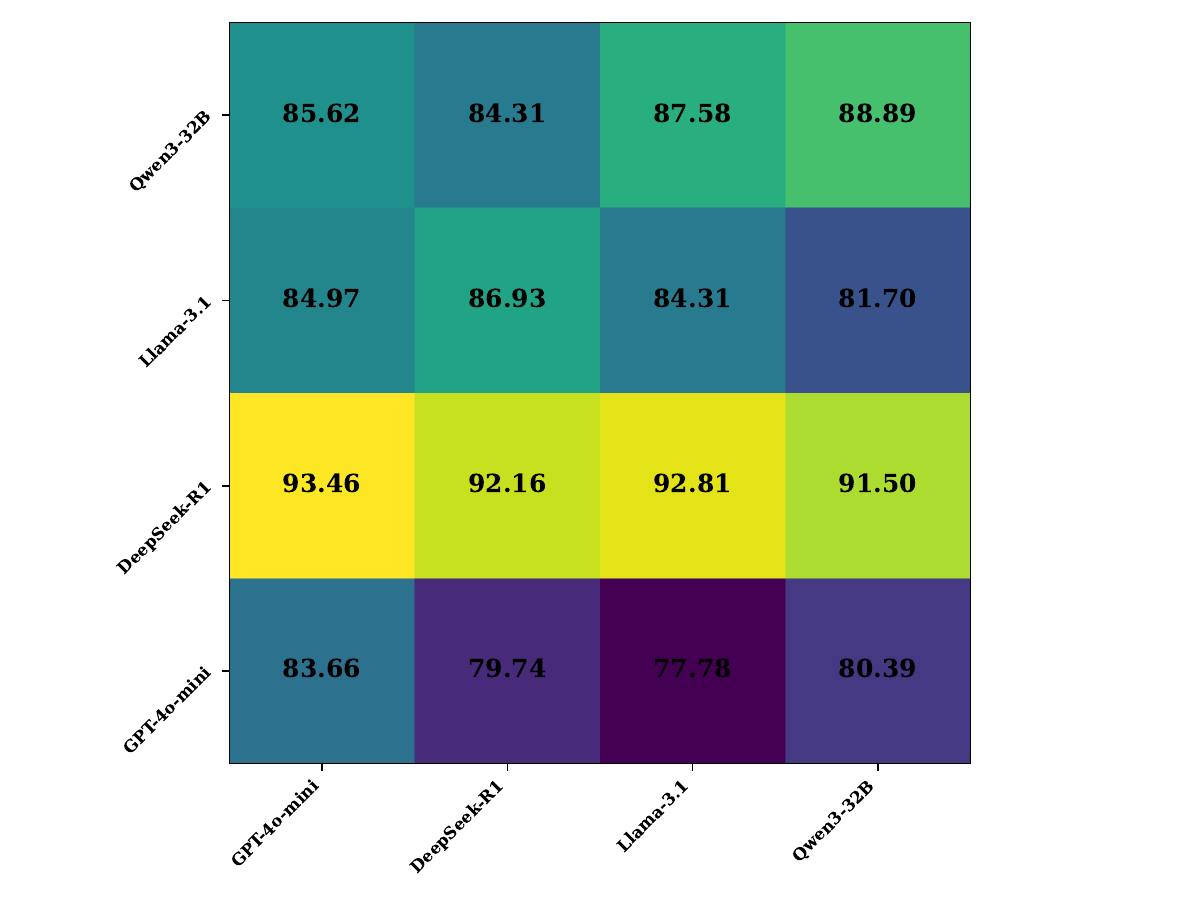} 
        \caption{LLM models on MMLU.}
        \label{fig:llm_model}
    \end{subfigure}
    \caption{Different number of agents and LLM models.}
    \label{fig:hyperparameter}
    %\vspace{-0.3in}
\end{figure}

\textbf{Impact of the hyper-parameters.} We analyze the influence of two factors: the number of agents and the choice of LLMs. As presented in Figure (\ref{fig:agent_num}), when the number of agents increases, performance initially improves; however, the marginal gains diminish beyond a certain point. For complex tasks like MultiArith, increasing the number of agents can lead to continued performance improvements. In contrast, for simpler tasks like MMLU, adding more agents may introduce unnecessary redundancy and even degrade performance. Figure (\ref{fig:llm_model}) further examines the transferability of the generative model across different LLM backbones on the MMLU (i.e., \texttt{gpt-4o-mini}, \texttt{Llama-3.1-70B Instruct}, \texttt{Qwen3-32B} and \texttt{DeepSeek-R1}). The \texttt{x-axis} shows the training LLM, and the \texttt{y-axis} shows the evaluation LLM. The results show that DeepSeek-R1 exhibits strong reasoning capabilities, even when the generative model is trained with a comparatively weaker LLM. Thus, our model enables training with weaker, more cost-effective LLMs while leveraging stronger, more expensive models during evaluation and deployment.

\textbf{Impact of multi-agent collaboration framework.} To verify the effectiveness of multi-agent collaboration over single-agent settings, we evaluate RADAR across a range of base models with varying capabilities, including stronger models such as DeepSeek-R1. The results are summarized in Table \ref{tab:mas}. We observe that RADAR consistently improves performance across all models, including stronger ones. While the absolute gains become smaller for more capable models (e.g., +1.35 for DeepSeek-R1 vs. larger gains for weaker models), the improvements remain stable. This suggests that RADAR provides complementary benefits beyond base model capability, such as structured collaboration and reduced redundancy, which remain useful even when single-agent performance is strong.

\begin{table}
    \centering
    \caption{Multi-agent collaboration vs. single-agent performance.}
    \begin{tabular}{lcc}
        \Xhline{1.0pt}
        \textbf{LLMs} & \textbf{Single Agents} & \textbf{RADAR(MAS)}   \\
        \Xhline{1.0pt}
        gpt-4o-min & 78.54 &  83.66$_{\textcolor{orange}{\uparrow 5.12}}$ \\
        Llama-3.1-70B & 77.12 & 84.31$_{\textcolor{orange}{\uparrow 7.19}}$  \\
        Qwen3-32B & 84.97 &  88.89$_{\textcolor{orange}{\uparrow 3.92}}$ \\
        DeepSeek-R1 & 90.81 &  92.16$_{\textcolor{orange}{\uparrow 1.35}}$  \\
        \Xhline{1.0pt}
    \end{tabular}
    \label{tab:mas}
    \vspace{-0.1in}
\end{table}

\textbf{Efficiency.} We provide a comparison of training time, inference time, and overall token consumption on GSM8K dataset below. As shown in Table \ref{tab:efficiency}, RADAR reduces token consumption, while incurring a moderate increase in inference time due to the additional graph generation process. Training time remains comparable to existing methods. Regarding inference time, AFlow is a search-based method that learns a single optimized workflow shared across all queries, making it faster at inference. In contrast, our method generates query-adaptive topologies for each instance. While this per-query adaptation introduces additional overhead, it enables more efficient communication tailored to task complexity.

\begin{table}
    \centering
    \caption{Comparison of wall-clock time and token consumption.}
    \begin{tabular}{lccc}
        \Xhline{1.0pt}
        \textbf{Methods} & \textbf{Overall Token} & \textbf{Train} & \textbf{Infer}   \\
        \Xhline{1.0pt}
        AFlow  & $1.4 \times10^7$ &  2h43min & 7.32min \\
        AgentPrune   & $1.1 \times 10^7$  &  1h27min  & 14.25min \\
        RADAR  & $6.5 \times 10^6$ &  2h10min  & 17.55min \\
        \Xhline{1.0pt}
    \end{tabular}
    \label{tab:efficiency}
\end{table}

\textbf{Distribution of Generated Collaboration Graphs Across Models.} We analyze the distribution of generated graph structures by reporting the mean and standard deviation of graph size, density, and effective size across methods on MMLU dataset. We note that although the number of agents is set to 5, some nodes may remain inactive due to isolation; we therefore report statistics over the active subgraph during execution. From these results in Table \ref{tab:distribution}, RADAR exhibits two key structural differences: (1) Higher effective size (lower redundancy). RADAR achieves significantly higher effective size (0.92 vs. 0.73 / 0.68), indicating that it produces less redundant and more informative communication structures. (2) More efficient connectivity. RADAR maintains comparable graph size and slightly lower density, suggesting that it avoids unnecessary connections while preserving sufficient coordination. In contrast, baseline methods tend to produce either denser or less structured topologies with lower effective size, leading to more redundant communication.

\begin{table}
    \centering
    \caption{Distribution of graph sizes (GS), densities, and effective sizes (ES). ARG is short for ARG-Designer.}
    \begin{tabular}{lccc}
        \Xhline{1.0pt}
        \textbf{Methods} & \textbf{GS} & \textbf{Densities} & \textbf{ES}   \\
        \Xhline{1.0pt}
        GDesigner  & 4.92$\pm$0.25 & 0.302$\pm$0.065 & 0.73$\pm$0.17 \\
        ARG   & 3.79$\pm$1.57  &  0.317$\pm$0.133  & 0.68$\pm$0.13 \\
        RADAR  & 4.60$\pm$0.66 &  0.289$\pm$0.071  & 0.92$\pm$0.18 \\
        \Xhline{1.0pt}
    \end{tabular}
    \label{tab:distribution}
    \vspace{-0.2in}
\end{table}

\section{Conclusion}
In this paper, we propose an iterative framework for multi-agent collaboration topology design, which provides high accuracy, low token consumption, and strong robustness across diverse scenarios. Specifically, we employ a graph diffusion model that leverages graph effectiveness signals to iteratively construct collaboration topologies, explicitly modeling redundancy as part of the topology generation process. The proposed framework is flexible and task-adaptive, enabling effective topology customization for different queries. We conduct extensive experiments on six benchmarks spanning general reasoning, code generation and mathematical problem solving, which collectively demonstrate the effectiveness of our method. Extending the framework to support dynamic role composition or generation is an interesting direction for future work.

\section*{Acknowledgements}
This work is sponsored by CCF-Tencent Rhino-Bird Open Research Fund and is supported by the ``111 Center" (No. B26023).

\section*{Impact Statement}
\textbf{Ethical Aspects.} We do not identify any ethical concerns arising from the motivation, methodology, experimental setup, or data usage in this work. The proposed RADAR framework is designed to advance research in multi-agent systems and automated communication topology design in a responsible manner, with the goal of improving efficiency and robustness in collaborative AI systems.

\textbf{Societal Consequences.} In this work, we introduce a new paradigm for multi-agent system design based on iterative, redundancy-aware topology generation. By enabling fine-grained and task-adaptive resource allocation, RADAR improves efficiency while maintaining high output quality. Reduced inference costs and increased flexibility in multi-agent workflows may help lower barriers to adopting intelligent automation, with potential benefits for a broad range of applications in education, scientific research, and industry.


\bibliography{reference}

@inproceedings{zhang2024codeagent,
  title={CodeAgent: Enhancing Code Generation with Tool-Integrated Agent Systems for Real-World Repo-level Coding Challenges},
  author={Zhang, Kechi and Li, Jia and Li, Ge and Shi, Xianjie and Jin, Zhi},
  booktitle={Proceedings of the 62nd Annual Meeting of the Association for Computational Linguistics (Volume 1: Long Papers)},
  pages={13643--13658},
  year={2024}
}

@inproceedings{xu2024generate,
  title={Generate-on-Graph: Treat LLM as both Agent and KG for Incomplete Knowledge Graph Question Answering},
  author={Xu, Yao and He, Shizhu and Chen, Jiabei and Wang, Zihao and Song, Yangqiu and Tong, Hanghang and Liu, Guang and Zhao, Jun and Liu, Kang},
  booktitle={Proceedings of the 2024 Conference on Empirical Methods in Natural Language Processing},
  pages={18410--18430},
  year={2024}
}

@inproceedings{chae2025web,
title={Web Agents with World Models: Learning and Leveraging Environment Dynamics in Web Navigation},
author={Hyungjoo Chae and Namyoung Kim and Kai Tzu-iunn Ong and Minju Gwak and Gwanwoo Song and Jihoon Kim and Sunghwan Kim and Dongha Lee and Jinyoung Yeo},
booktitle={The Thirteenth International Conference on Learning Representations},
year={2025},
url={https://openreview.net/forum?id=moWiYJuSGF}
}

@inproceedings{zhu2025multiagentbench,
  title={Multiagentbench: Evaluating the collaboration and competition of llm agents},
  author={Zhu, Kunlun and Du, Hongyi and Hong, Zhaochen and Yang, Xiaocheng and Guo, Shuyi and Wang, Daisy Zhe and Wang, Zhenhailong and Qian, Cheng and Tang, Robert and Ji, Heng and others},
  booktitle={Proceedings of the 63rd Annual Meeting of the Association for Computational Linguistics (Volume 1: Long Papers)},
  pages={8580--8622},
  year={2025}
}

@inproceedings{wu2024shall,
  title={Shall we team up: Exploring spontaneous cooperation of competing llm agents},
  author={Wu, Zengqing and Peng, Run and Zheng, Shuyuan and Liu, Qianying and Han, Xu and Kwon, Brian I and Onizuka, Makoto and Tang, Shaojie and Xiao, Chuan},
  booktitle={Findings of the Association for Computational Linguistics: EMNLP 2024},
  pages={5163--5186},
  year={2024}
}

@inproceedings{zhang2025cut,
title={Cut the Crap: An Economical Communication Pipeline for {LLM}-based Multi-Agent Systems},
author={Guibin Zhang and Yanwei Yue and Zhixun Li and Sukwon Yun and Guancheng Wan and Kun Wang and Dawei Cheng and Jeffrey Xu Yu and Tianlong Chen},
booktitle={The Thirteenth International Conference on Learning Representations},
year={2025},
url={https://openreview.net/forum?id=LkzuPorQ5L}
}

@inproceedings{piatti2024cooperate,
title={Cooperate or Collapse:  Emergence of Sustainable Cooperation in a Society of {LLM} Agents},
author={Giorgio Piatti and Zhijing Jin and Max Kleiman-Weiner and Bernhard Sch{\"o}lkopf and Mrinmaya Sachan and Rada Mihalcea},
booktitle={The Thirty-eighth Annual Conference on Neural Information Processing Systems},
year={2024},
url={https://openreview.net/forum?id=0zWzJj6lO3}
}

@inproceedings{chen2025debatecoder,
  title={DebateCoder: Towards Collective Intelligence of LLMs via Test Case Driven LLM Debate for Code Generation},
  author={Chen, Jizheng and Du, Kounianhua and Dai, Xinyi and Zhang, Weiming and Wang, Xihuai and Wang, Yasheng and Tang, Ruiming and Zhang, Weinan and Yu, Yong},
  booktitle={Proceedings of the 63rd Annual Meeting of the Association for Computational Linguistics (Volume 1: Long Papers)},
  pages={12055--12065},
  year={2025}
}

@article{jin2025controlling,
  title={Controlling Performance and Budget of a Centralized Multi-agent LLM System with Reinforcement Learning},
  author={Jin, Bowen and Collins, TJ and Yu, Donghan and Cemri, Mert and Zhang, Shenao and Li, Mengyu and Tang, Jay and Qin, Tian and Xu, Zhiyang and Lu, Jiarui and others},
  journal={arXiv preprint arXiv:2511.02755},
  year={2025}
}

@article{wei2022chain,
  title={Chain-of-thought prompting elicits reasoning in large language models},
  author={Wei, Jason and Wang, Xuezhi and Schuurmans, Dale and Bosma, Maarten and Xia, Fei and Chi, Ed and Le, Quoc V and Zhou, Denny and others},
  journal={Advances in neural information processing systems},
  volume={35},
  pages={24824--24837},
  year={2022}
}

@article{wang2025agentdropout,
  title={Agentdropout: Dynamic agent elimination for token-efficient and high-performance llm-based multi-agent collaboration},
  author={Wang, Zhexuan and Wang, Yutong and Liu, Xuebo and Ding, Liang and Zhang, Miao and Liu, Jie and Zhang, Min},
  journal={arXiv preprint arXiv:2503.18891},
  year={2025}
}

@inproceedings{kong2023autoregressive,
  title={Autoregressive diffusion model for graph generation},
  author={Kong, Lingkai and Cui, Jiaming and Sun, Haotian and Zhuang, Yuchen and Prakash, B Aditya and Zhang, Chao},
  booktitle={International conference on machine learning},
  pages={17391--17408},
  year={2023},
  organization={PMLR}
}

@book{ronald1992structural,
 author={Ronald S. Burt},
 publisher={Harvard University Press},
 title={Structural Holes: The Social Structure of Competition},
 year={1992}
}

@inproceedings{shen2023hugginggpt,
title={Hugging{GPT}: Solving {AI} Tasks with Chat{GPT} and its Friends in Hugging Face},
author={Yongliang Shen and Kaitao Song and Xu Tan and Dongsheng Li and Weiming Lu and Yueting Zhuang},
booktitle={Thirty-seventh Conference on Neural Information Processing Systems},
year={2023},
url={https://openreview.net/forum?id=yHdTscY6Ci}
}

@inproceedings{wang2024rethinking,
  title={Rethinking the Bounds of LLM Reasoning: Are Multi-Agent Discussions the Key?},
  author={Wang, Qineng and Wang, Zihao and Su, Ying and Tong, Hanghang and Song, Yangqiu},
  booktitle={Proceedings of the 62nd Annual Meeting of the Association for Computational Linguistics (Volume 1: Long Papers)},
  pages={6106--6131},
  year={2024}
}

@inproceedings{song2023llm,
  title={Llm-planner: Few-shot grounded planning for embodied agents with large language models},
  author={Song, Chan Hee and Wu, Jiaman and Washington, Clayton and Sadler, Brian M and Chao, Wei-Lun and Su, Yu},
  booktitle={Proceedings of the IEEE/CVF international conference on computer vision},
  pages={2998--3009},
  year={2023}
}

@inproceedings{hong2024metagpt,
title={Meta{GPT}: Meta Programming for A Multi-Agent Collaborative Framework},
author={Sirui Hong and Mingchen Zhuge and Jonathan Chen and Xiawu Zheng and Yuheng Cheng and Jinlin Wang and Ceyao Zhang and Zili Wang and Steven Ka Shing Yau and Zijuan Lin and Liyang Zhou and Chenyu Ran and Lingfeng Xiao and Chenglin Wu and J{\"u}rgen Schmidhuber},
booktitle={The Twelfth International Conference on Learning Representations},
year={2024},
url={https://openreview.net/forum?id=VtmBAGCN7o}
}

@inproceedings{wu2024autogen,
title={AutoGen: Enabling Next-Gen {LLM} Applications via Multi-Agent Conversations},
author={Qingyun Wu and Gagan Bansal and Jieyu Zhang and Yiran Wu and Beibin Li and Erkang Zhu and Li Jiang and Xiaoyun Zhang and Shaokun Zhang and Jiale Liu and Ahmed Hassan Awadallah and Ryen W White and Doug Burger and Chi Wang},
booktitle={First Conference on Language Modeling},
year={2024},
url={https://openreview.net/forum?id=BAakY1hNKS}
}

@inproceedings{chen2023efficient,
  title={Efficient and Degree-Guided Graph Generation via Discrete Diffusion Modeling},
  author={Chen, Xiaohui and He, Jiaxing and Han, Xu and Liu, Liping},
  booktitle={International Conference on Machine Learning},
  pages={4585--4610},
  year={2023},
  organization={PMLR}
}

@article{liu2024graph,
  title={Graph diffusion transformers for multi-conditional molecular generation},
  author={Liu, Gang and Xu, Jiaxin and Luo, Tengfei and Jiang, Meng},
  journal={Advances in Neural Information Processing Systems},
  volume={37},
  pages={8065--8092},
  year={2024}
}

@article{yi2023graph,
  title={Graph denoising diffusion for inverse protein folding},
  author={Yi, Kai and Zhou, Bingxin and Shen, Yiqing and Li{\`o}, Pietro and Wang, Yuguang},
  journal={Advances in Neural Information Processing Systems},
  volume={36},
  pages={10238--10257},
  year={2023}
}

@inproceedings{wu2025agentic,
  title={Agentic reasoning: A streamlined framework for enhancing llm reasoning with agentic tools},
  author={Wu, Junde and Zhu, Jiayuan and Liu, Yuyuan and Xu, Min and Jin, Yueming},
  booktitle={Proceedings of the 63rd Annual Meeting of the Association for Computational Linguistics (Volume 1: Long Papers)},
  pages={28489--28503},
  year={2025}
}

@inproceedings{kipf2017semisupervised,
  title={Semi-Supervised Classification with Graph Convolutional Networks},
  author={Thomas N. Kipf and Max Welling},
  booktitle={International Conference on Learning Representations},
  year={2017},
  url={https://openreview.net/forum?id=SJU4ayYgl}
}

@article{hamilton2017inductive,
  title={Inductive representation learning on large graphs},
  author={Hamilton, Will and Ying, Zhitao and Leskovec, Jure},
  journal={Advances in neural information processing systems},
  volume={30},
  year={2017}
}

@inproceedings{veličković2018graph,
title={Graph Attention Networks},
author={Petar Veličković and Guillem Cucurull and Arantxa Casanova and Adriana Romero and Pietro Liò and Yoshua Bengio},
booktitle={International Conference on Learning Representations},
year={2018},
url={https://openreview.net/forum?id=rJXMpikCZ},
}

@inproceedings{zhang2021h2mn,
  title={H2mn: Graph similarity learning with hierarchical hypergraph matching networks},
  author={Zhang, Zhen and Bu, Jiajun and Ester, Martin and Li, Zhao and Yao, Chengwei and Yu, Zhi and Wang, Can},
  booktitle={Proceedings of the 27th ACM SIGKDD conference on knowledge discovery \& data mining},
  pages={2274--2284},
  year={2021}
}

@inproceedings{li2023api,
  title={API-Bank: A Comprehensive Benchmark for Tool-Augmented LLMs},
  author={Li, Minghao and Zhao, Yingxiu and Yu, Bowen and Song, Feifan and Li, Hangyu and Yu, Haiyang and Li, Zhoujun and Huang, Fei and Li, Yongbin},
  booktitle={Proceedings of the 2023 Conference on Empirical Methods in Natural Language Processing},
  pages={3102--3116},
  year={2023}
}

@inproceedings{
hendrycks2021measuring,
title={Measuring Massive Multitask Language Understanding},
author={Dan Hendrycks and Collin Burns and Steven Basart and Andy Zou and Mantas Mazeika and Dawn Song and Jacob Steinhardt},
booktitle={International Conference on Learning Representations},
year={2021},
url={https://openreview.net/forum?id=d7KBjmI3GmQ}
}

@article{cobbe2021training,
  title={Training verifiers to solve math word problems},
  author={Cobbe, Karl and Kosaraju, Vineet and Bavarian, Mohammad and Chen, Mark and Jun, Heewoo and Kaiser, Lukasz and Plappert, Matthias and Tworek, Jerry and Hilton, Jacob and Nakano, Reiichiro and others},
  journal={arXiv preprint arXiv:2110.14168},
  year={2021}
}

@inproceedings{roy2015solving,
  title={Solving general arithmetic word problems},
  author={Roy, Subhro and Roth, Dan},
  booktitle={Proceedings of the 2015 conference on empirical methods in natural language processing},
  pages={1743--1752},
  year={2015}
}

@inproceedings{patel2021nlp,
  title={Are NLP Models really able to Solve Simple Math Word Problems?},
  author={Patel, Arkil and Bhattamishra, Satwik and Goyal, Navin},
  booktitle={Proceedings of the 2021 Conference of the North American Chapter of the Association for Computational Linguistics: Human Language Technologies},
  pages={2080--2094},
  year={2021}
}

@misc{chen2021evaluatinglargelanguagemodels,
      title={Evaluating Large Language Models Trained on Code}, 
      author={Mark Chen and Jerry Tworek and Heewoo Jun and Qiming Yuan and Henrique Ponde de Oliveira Pinto and Jared Kaplan and Harri Edwards and Yuri Burda and Nicholas Joseph and Greg Brockman and Alex Ray and Raul Puri and Gretchen Krueger and Michael Petrov and Heidy Khlaaf and Girish Sastry and Pamela Mishkin and Brooke Chan and Scott Gray and Nick Ryder and Mikhail Pavlov and Alethea Power and Lukasz Kaiser and Mohammad Bavarian and Clemens Winter and Philippe Tillet and Felipe Petroski Such and Dave Cummings and Matthias Plappert and Fotios Chantzis and Elizabeth Barnes and Ariel Herbert-Voss and William Hebgen Guss and Alex Nichol and Alex Paino and Nikolas Tezak and Jie Tang and Igor Babuschkin and Suchir Balaji and Shantanu Jain and William Saunders and Christopher Hesse and Andrew N. Carr and Jan Leike and Josh Achiam and Vedant Misra and Evan Morikawa and Alec Radford and Matthew Knight and Miles Brundage and Mira Murati and Katie Mayer and Peter Welinder and Bob McGrew and Dario Amodei and Sam McCandlish and Ilya Sutskever and Wojciech Zaremba},
      year={2021},
      eprint={2107.03374},
      archivePrefix={arXiv},
      primaryClass={cs.LG},
      url={https://arxiv.org/abs/2107.03374}, 
}

@article{wang2020minilm,
  title={Minilm: Deep self-attention distillation for task-agnostic compression of pre-trained transformers},
  author={Wang, Wenhui and Wei, Furu and Dong, Li and Bao, Hangbo and Yang, Nan and Zhou, Ming},
  journal={Advances in neural information processing systems},
  volume={33},
  pages={5776--5788},
  year={2020}
}

@article{liu2024revisiting,
  title={Revisiting, benchmarking and understanding unsupervised graph domain adaptation},
  author={Liu, Meihan and Zhang, Zhen and Tang, Jiachen and Bu, Jiajun and He, Bingsheng and Zhou, Sheng},
  journal={Advances in neural information processing systems},
  volume={37},
  pages={89408--89436},
  year={2024}
}

@article{kingma2014adam,
  title={Adam: A method for stochastic optimization},
  author={Kingma, Diederik P},
  journal={arXiv preprint arXiv:1412.6980},
  year={2014}
}

@inproceedings{schlichtkrull2018modeling,
  title={Modeling relational data with graph convolutional networks},
  author={Schlichtkrull, Michael and Kipf, Thomas N and Bloem, Peter and Van Den Berg, Rianne and Titov, Ivan and Welling, Max},
  booktitle={European semantic web conference},
  pages={593--607},
  year={2018},
  organization={Springer}
}

@inproceedings{ling2017program,
  title={Program Induction by Rationale Generation: Learning to Solve and Explain Algebraic Word Problems},
  author={Ling, Wang and Yogatama, Dani and Dyer, Chris and Blunsom, Phil},
  booktitle={Proceedings of the 55th Annual Meeting of the Association for Computational Linguistics (Volume 1: Long Papers)},
  pages={158--167},
  year={2017}
}

@article{williams1992simple,
  title={Simple statistical gradient-following algorithms for connectionist reinforcement learning},
  author={Williams, Ronald J},
  journal={Machine learning},
  volume={8},
  number={3},
  pages={229--256},
  year={1992},
  publisher={Springer}
}

@article{vaswani2017attention,
  title={Attention is all you need},
  author={Vaswani, Ashish and Shazeer, Noam and Parmar, Niki and Uszkoreit, Jakob and Jones, Llion and Gomez, Aidan N and Kaiser, {\L}ukasz and Polosukhin, Illia},
  journal={Advances in neural information processing systems},
  volume={30},
  year={2017}
}

@article{yang2023directional,
  title={Directional diffusion models for graph representation learning},
  author={Yang, Run and Yang, Yuling and Zhou, Fan and Sun, Qiang},
  journal={Advances in Neural Information Processing Systems},
  volume={36},
  pages={32720--32731},
  year={2023}
}

@inproceedings{zhang2025towards,
title={Towards Unsupervised Open-Set Graph Domain Adaptation via Dual Reprogramming},
author={Zhen Zhang and Bingsheng He},
booktitle={The Thirty-ninth Annual Conference on Neural Information Processing Systems},
year={2025},
url={https://openreview.net/forum?id=iDcPkDrlaW}
}

@article{chen2024more,
  title={Are more llm calls all you need? towards the scaling properties of compound ai systems},
  author={Chen, Lingjiao and Davis, Jared and Hanin, Boris and Bailis, Peter and Stoica, Ion and Zaharia, Matei and Zou, James},
  journal={Advances in Neural Information Processing Systems},
  volume={37},
  pages={45767--45790},
  year={2024}
}

@inproceedings{wolflein2025llm,
  title={Llm agents making agent tools},
  author={W{\"o}lflein, Georg and Ferber, Dyke and Truhn, Daniel and Arandjelovic, Ognjen and Kather, Jakob Nikolas},
  booktitle={Proceedings of the 63rd Annual Meeting of the Association for Computational Linguistics (Volume 1: Long Papers)},
  pages={26092--26130},
  year={2025}
}

@inproceedings{yuan2025evoagent,
  title={Evoagent: Towards automatic multi-agent generation via evolutionary algorithms},
  author={Yuan, Siyu and Song, Kaitao and Chen, Jiangjie and Tan, Xu and Li, Dongsheng and Yang, Deqing},
  booktitle={Proceedings of the 2025 Conference of the Nations of the Americas Chapter of the Association for Computational Linguistics: Human Language Technologies (Volume 1: Long Papers)},
  pages={6192--6217},
  year={2025}
}

@inproceedings{
zhang2023automatic,
title={Automatic Chain of Thought Prompting in Large Language Models},
author={Zhuosheng Zhang and Aston Zhang and Mu Li and Alex Smola},
booktitle={The Eleventh International Conference on Learning Representations },
year={2023},
url={https://openreview.net/forum?id=5NTt8GFjUHkr}
}

@article{yao2023tree,
  title={Tree of thoughts: Deliberate problem solving with large language models},
  author={Yao, Shunyu and Yu, Dian and Zhao, Jeffrey and Shafran, Izhak and Griffiths, Tom and Cao, Yuan and Narasimhan, Karthik},
  journal={Advances in neural information processing systems},
  volume={36},
  pages={11809--11822},
  year={2023}
}

@inproceedings{fu2023complexitybased,
title={Complexity-Based Prompting for Multi-step Reasoning},
author={Yao Fu and Hao Peng and Ashish Sabharwal and Peter Clark and Tushar Khot},
booktitle={The Eleventh International Conference on Learning Representations },
year={2023},
url={https://openreview.net/forum?id=yf1icZHC-l9}
}

@inproceedings{wang2023selfconsistency,
title={Self-Consistency Improves Chain of Thought Reasoning in Language Models},
author={Xuezhi Wang and Jason Wei and Dale Schuurmans and Quoc V Le and Ed H. Chi and Sharan Narang and Aakanksha Chowdhery and Denny Zhou},
booktitle={The Eleventh International Conference on Learning Representations },
year={2023},
url={https://openreview.net/forum?id=1PL1NIMMrw}
}

@inproceedings{wang2024unleashing,
  title={Unleashing the emergent cognitive synergy in large language models: A task-solving agent through multi-persona self-collaboration},
  author={Wang, Zhenhailong and Mao, Shaoguang and Wu, Wenshan and Ge, Tao and Wei, Furu and Ji, Heng},
  booktitle={Proceedings of the 2024 Conference of the North American Chapter of the Association for Computational Linguistics: Human Language Technologies (Volume 1: Long Papers)},
  pages={257--279},
  year={2024}
}

@inproceedings{du2023improving,
  title={Improving factuality and reasoning in language models through multiagent debate},
  author={Du, Yilun and Li, Shuang and Torralba, Antonio and Tenenbaum, Joshua B and Mordatch, Igor},
  booktitle={Forty-first International Conference on Machine Learning},
  year={2023}
}

@inproceedings{jiang2023llm,
  title={LLM-Blender: Ensembling Large Language Models with Pairwise Ranking and Generative Fusion},
  author={Jiang, Dongfu and Ren, Xiang and Lin, Bill Yuchen},
  booktitle={Proceedings of the 61st Annual Meeting of the Association for Computational Linguistics (Volume 1: Long Papers)},
  pages={14165--14178},
  year={2023}
}

@inproceedings{liu2024dynamic,
  title={A dynamic llm-powered agent network for task-oriented agent collaboration},
  author={Liu, Zijun and Zhang, Yanzhe and Li, Peng and Liu, Yang and Yang, Diyi},
  booktitle={First Conference on Language Modeling},
  year={2024}
}

@inproceedings{chen2024agentverse,
title={AgentVerse: Facilitating Multi-Agent Collaboration and Exploring Emergent Behaviors},
author={Weize Chen and Yusheng Su and Jingwei Zuo and Cheng Yang and Chenfei Yuan and Chi-Min Chan and Heyang Yu and Yaxi Lu and Yi-Hsin Hung and Chen Qian and Yujia Qin and Xin Cong and Ruobing Xie and Zhiyuan Liu and Maosong Sun and Jie Zhou},
booktitle={The Twelfth International Conference on Learning Representations},
year={2024},
url={https://openreview.net/forum?id=EHg5GDnyq1}
}

@inproceedings{qian2025scaling,
title={Scaling Large Language Model-based Multi-Agent Collaboration},
author={Chen Qian and Zihao Xie and YiFei Wang and Wei Liu and Kunlun Zhu and Hanchen Xia and Yufan Dang and Zhuoyun Du and Weize Chen and Cheng Yang and Zhiyuan Liu and Maosong Sun},
booktitle={The Thirteenth International Conference on Learning Representations},
year={2025},
url={https://openreview.net/forum?id=K3n5jPkrU6}
}

@inproceedings{chen2024autoagents,
  title={AutoAgents: a framework for automatic agent generation},
  author={Chen, Guangyao and Dong, Siwei and Shu, Yu and Zhang, Ge and Sesay, Jaward and Karlsson, B{\"o}rje and Fu, Jie and Shi, Yemin},
  booktitle={Proceedings of the Thirty-Third International Joint Conference on Artificial Intelligence},
  pages={22--30},
  year={2024}
}

@inproceedings{zhuge2024gptswarm,
title={{GPTS}warm: Language Agents as Optimizable Graphs},
author={Mingchen Zhuge and Wenyi Wang and Louis Kirsch and Francesco Faccio and Dmitrii Khizbullin and J{\"u}rgen Schmidhuber},
booktitle={Forty-first International Conference on Machine Learning},
year={2024},
url={https://openreview.net/forum?id=uTC9AFXIhg}
}

@inproceedings{hu2025automated,
title={Automated Design of Agentic Systems},
author={Shengran Hu and Cong Lu and Jeff Clune},
booktitle={The Thirteenth International Conference on Learning Representations},
year={2025},
url={https://openreview.net/forum?id=t9U3LW7JVX}
}

@inproceedings{shang2025agentsquare,
title={AgentSquare: Automatic {LLM} Agent Search in Modular Design Space},
author={Yu Shang and Yu Li and Keyu Zhao and Likai Ma and Jiahe Liu and Fengli Xu and Yong Li},
booktitle={The Thirteenth International Conference on Learning Representations},
year={2025},
url={https://openreview.net/forum?id=mPdmDYIQ7f}
}

@inproceedings{zhang2025aflow,
title={{AF}low: Automating Agentic Workflow Generation},
author={Jiayi Zhang and Jinyu Xiang and Zhaoyang Yu and Fengwei Teng and Xiong-Hui Chen and Jiaqi Chen and Mingchen Zhuge and Xin Cheng and Sirui Hong and Jinlin Wang and Bingnan Zheng and Bang Liu and Yuyu Luo and Chenglin Wu},
booktitle={The Thirteenth International Conference on Learning Representations},
year={2025},
url={https://openreview.net/forum?id=z5uVAKwmjf}
}

@inproceedings{zhang2025gdesigner,
title={G-Designer: Architecting Multi-agent Communication Topologies via Graph Neural Networks},
author={Guibin Zhang and Yanwei Yue and Xiangguo Sun and Guancheng Wan and Miao Yu and Junfeng Fang and Kun Wang and Tianlong Chen and Dawei Cheng},
booktitle={Forty-second International Conference on Machine Learning},
year={2025},
url={https://openreview.net/forum?id=LpE54NUnmO}
}

@inproceedings{zhang2025multiagent,
title={Multi-agent Architecture Search via Agentic Supernet},
author={Guibin Zhang and Luyang Niu and Junfeng Fang and Kun Wang and LEI BAI and Xiang Wang},
booktitle={Forty-second International Conference on Machine Learning},
year={2025},
url={https://openreview.net/forum?id=imcyVlzpXh}
}

@article{li2025assemble,
  title={Assemble your crew: Automatic multi-agent communication topology design via autoregressive graph generation},
  author={Li, Shiyuan and Liu, Yixin and Wen, Qingsong and Zhang, Chengqi and Pan, Shirui},
  journal={arXiv preprint arXiv:2507.18224},
  year={2025}
}

@article{jiang2025dynamic,
  title={Dynamic Generation of Multi-LLM Agents Communication Topologies with Graph Diffusion Models},
  author={Jiang, Eric Hanchen and Wan, Guancheng and Yin, Sophia and Li, Mengting and Wu, Yuchen and Liang, Xiao and Li, Xinfeng and Sun, Yizhou and Wang, Wei and Chang, Kai-Wei and others},
  journal={arXiv preprint arXiv:2510.07799},
  year={2025}
}

@inproceedings{klipfel2024vector,
  title={Vector field oriented diffusion model for crystal material generation},
  author={Klipfel, Astrid and Fregier, Ya{\"e}l and Sayede, Adlane and Bouraoui, Zied},
  booktitle={Proceedings of the AAAI Conference on Artificial Intelligence},
  pages={22193--22201},
  year={2024}
}
\bibliographystyle{icml2026}

%%%%%%%%%%%%%%%%%%%%%%%%%%%%%%%%%%%%%%%%%%%%%%%%%%%%%%%%%%%%%%%%%%%%%%%%%%%%%%%
%%%%%%%%%%%%%%%%%%%%%%%%%%%%%%%%%%%%%%%%%%%%%%%%%%%%%%%%%%%%%%%%%%%%%%%%%%%%%%%
% APPENDIX
%%%%%%%%%%%%%%%%%%%%%%%%%%%%%%%%%%%%%%%%%%%%%%%%%%%%%%%%%%%%%%%%%%%%%%%%%%%%%%%
%%%%%%%%%%%%%%%%%%%%%%%%%%%%%%%%%%%%%%%%%%%%%%%%%%%%%%%%%%%%%%%%%%%%%%%%%%%%%%%
\newpage
\appendix
\onecolumn
\section{Dataset Statistics}
\label{ap:dataset}

\begin{table}
    \centering
    \caption{Dataset statistics.}
    \begin{tabular}{ccccc}
        \Xhline{1.0pt}
        \textbf{Category} & \textbf{Dataset} & \textbf{Format} & \textbf{Metric} & \textbf{\#Samples}  \\
        \Xhline{1.0pt}
         General Reasoning & MMLU & Multi-choice & Accuracy & 153 \\
         \Xhline{0.6pt}
         \multirow{4}{*}{Math Solving}& GSM8K & Number & Accuracy & 1,319 \\
         & MultiArith & Number & Accuracy & 600 \\
         & SVAMP & Number & Accuracy & 1,000 \\
         & AQuA & Multi-choice & Accuracy & 254 \\
        \Xhline{0.6pt}
        Code Generation & HumanEval & Code & Pass@1 & 164  \\
        \Xhline{1.0pt}
    \end{tabular}
    \label{tab:datasets}
\end{table}

We present the statistics of three types of datasets in Table \ref{tab:datasets}.

\section{Baselines}
\label{ap:baselines}
The configurations used for each baseline method are described in below:
\begin{itemize}
    \item \textbf{CoT \cite{wei2022chain}.} Chain-of-Thought (CoT) prompting enables LLM agents to perform step-by-step reasoning instead of generating direct answers. We follow the implementation described in \cite{zhang2023automatic}.
    \item \textbf{ComplexCoT \cite{fu2023complexitybased}.} Our experiments are based on the official implementation released at \url{https://github.com/FranxYao/Complexity-Based-Prompting/tree/main}.
    \item \textbf{SC(COT$\times$5) \cite{wang2023selfconsistency}.} For robustness, we aggregate five solutions produced using Chain-of-Thought prompting.
    \item \textbf{MultiPersona \cite{wang2024unleashing}.} It converts a single LLM into a cognitive synergist through multi-turn self-collaboration with multiple personas. We utilize the official code available at \url{https://github.com/MikeWangWZHL/Solo-Performance-Prompting}.
    \item \textbf{LLM-Debate \cite{du2023improving}.} We employ five role-specialized LLM agents that participate in up to two debate rounds, with the final output selected by majority voting. The implementation is based on \url{https://github.com/ucl-dark/llm_debate}.
    \item \textbf{LLM-Blender \cite{jiang2023llm}.} In experiments, LLM-Blender is instantiated using two \texttt{gpt-4o-mini} models, one \texttt{Qwen-2.5-72B}, and one \texttt{LLaMA-3.1-70B}. The source code is available at \url{https://github.com/yuchenlin/LLM-Blender}.
    \item \textbf{DyLAN \cite{liu2024dynamic}.} Our experiments are conducted using the implementation available at \url{https://github.com/SALT-NLP/DyLAN}.
    \item \textbf{AgentVerse \cite{chen2024agentverse}.} We follow the original implementation from \url{https://github.com/OpenBMB/AgentVerse}.
    \item \textbf{MacNet \cite{qian2025scaling}.} We use the ``MacNet-MESH" variant of MacNet, which employs a fully connected network topology. The implementation is available at \url{https://github.com/OpenBMB/ChatDev/tree/macnet}. 
    \item \textbf{AutoAgents \cite{chen2024autoagents}.} We adopt the configuration detailed in \url{https://github.com/Link-AGI/AutoAgents}.
    \item \textbf{GPTSwarm \cite{zhuge2024gptswarm}.} The method is executed using the original configuration as specified in \url{https://github.com/metauto-ai/GPTSwarm}.
    \item \textbf{ADAS \cite{hu2025automated}.} Our implementation is based on the authors' released code in \url{https://github.com/ShengranHu/ADAS}.
    \item \textbf{AgentSquare \cite{shang2025agentsquare}.} We build on the modular search framework from \cite{shang2025agentsquare}, using \texttt{GPT-4o-mini} as the fixed base LLM. Early stopping is applied, with training halted after five iterations without improvement. The codes are available at \url{https://github.com/tsinghua-fib-lab/AgentSquare}.
    \item \textbf{AFlow \cite{zhang2025aflow}.} To preserve experimental fairness under identical conditions, we configure AFlow to use \texttt{GPT-4o-mini} and limit the number of iterations to 20. We use the source code released at \url{https://github.com/FoundationAgents/AFlow}.
    \item \textbf{G-Designer \cite{zhang2025gdesigner}.} It employs a variational graph auto-encoder to construct the communication topology. We utilize the source code released at \url{https://github.com/yanweiyue/GDesigner}.
    \item \textbf{AgentPrune \cite{zhang2025cut}.} It performs pruning on the spatiotemporal message-passing graph. We utilize the source code available at \url{https://github.com/yanweiyue/AgentPrune}.
    \item \textbf{GTD \cite{jiang2025dynamic}.} We implement the method according to the original configuration specified in the paper.
    \item \textbf{MaAS \cite{zhang2025multiagent}.} It samples query-dependent agentic systems from the supernet. We use the source code available at \url{https://github.com/bingreeky/MaAS/}.
    \item \textbf{ARG-Designer \cite{li2025assemble}.} It dynamically determines the number of agents, assigns roles from an extensible pool, and establishes optimal communication links. We use the implementation available at \url{https://github.com/Shiy-Li/ARG-Designer}.
\end{itemize}

\section{More Ablation Studies}
\begin{figure}
    \centering
    \begin{subfigure}{0.3\linewidth}
        \includegraphics[width=\linewidth]{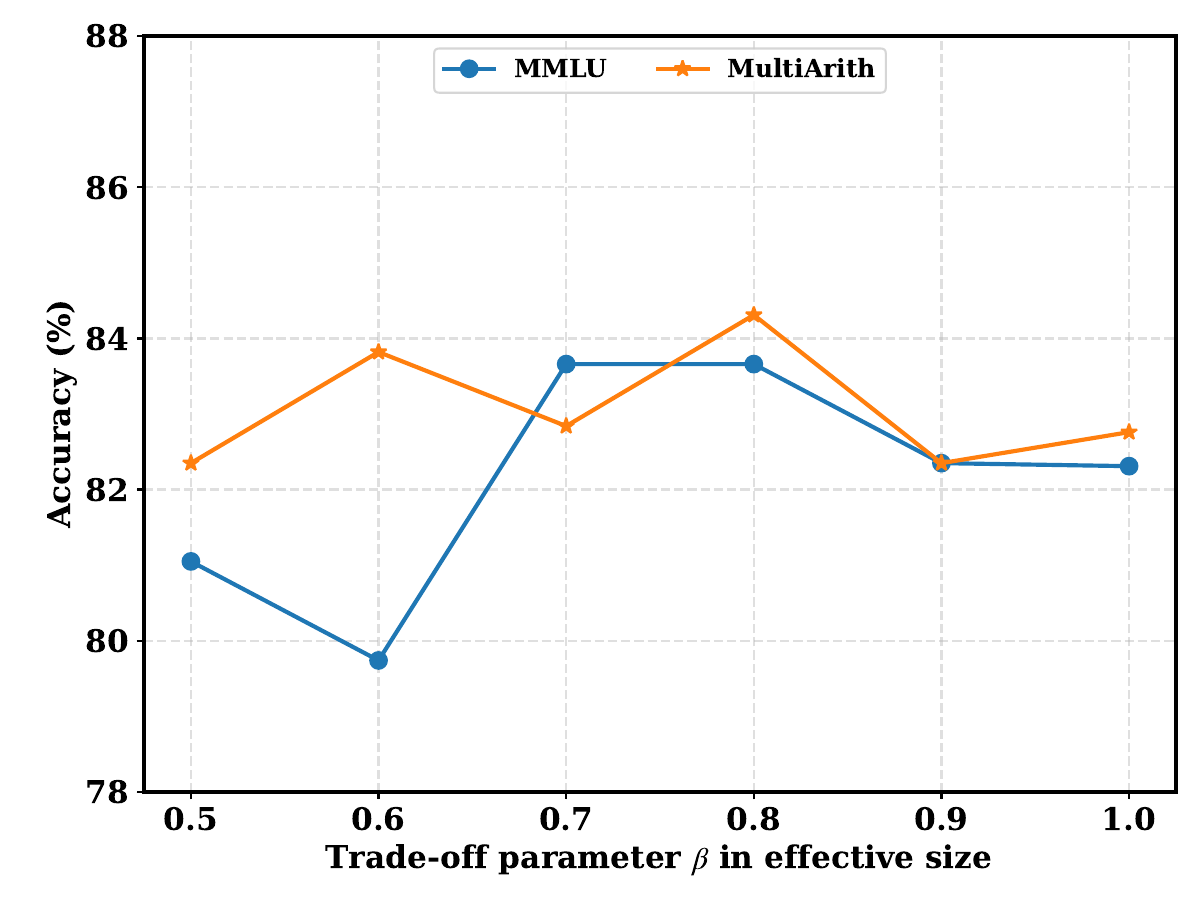} 
        \caption{Impact of trade-off parameter $\beta$.}
    \end{subfigure}
    \begin{subfigure}{0.3\linewidth}
        \includegraphics[width=\linewidth]{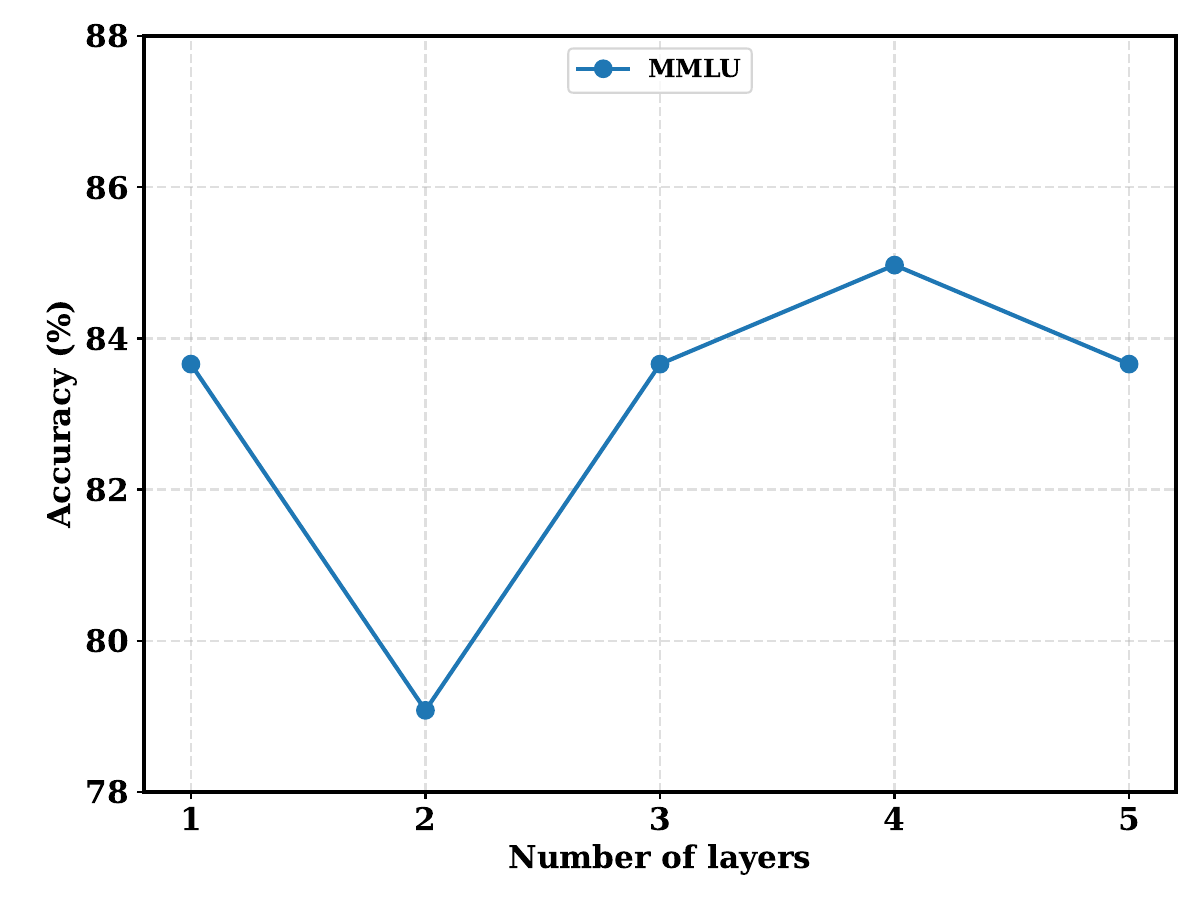} 
        \caption{Number of layers.}
    \end{subfigure}
    \caption{Hyper-parameter sensitivity analysis.}
    \label{fig:ab_parameter}
\end{figure}

We present additional ablation studies in Figure \ref{fig:ab_parameter}. Regarding the trade-off parameter $\beta$, which balances outgoing and incoming effective size, we assign greater weight to outgoing effective size. Outgoing redundancy affects multiple downstream agents simultaneously and thus dominates system-level cost, whereas incoming redundancy primarily impacts a single agent. Based on this analysis, we consistently set $\beta$ to either $0.7$ or $0.8$ to achieve a favorable performance. We also analyze the impact of the number of layers in the denoising network. The performance generally improves as the number of layers increases, but begins to decline beyond a certain point, indicating diminishing returns from overly complex architectures. Since generating excessively complicated structures is unnecessary, we set the number of layers to 3 or 4 for most datasets.

We also explored alternative GNN architectures for both ordering network and denoising network. Specifically, we replaced the RGCN in the ordering network with RGAT, and the GAT in the denoising network with GCN. The results in Table \ref{tab:backbone} indicate that these alternatives achieve comparable but generally slightly worse performance than our default design. This suggests that while the overall framework is not highly sensitive to the specific GNN choice, the selected combination (RGCN for ordering and GAT for denoising) provides a better balance between relational modeling and attention-based refinement.

Meanwhile, we extended our robustness evaluation to include both structural attacks and combined prompt + structural attacks, in addition to the original prompt-based setting. For prompt attacks, we simulate system prompt perturbations by converting two of the five agents into liar agents. For structural attacks, we inject noise into the collaboration graph by randomly adding 50\% additional edges. As shown in the Table \ref{tab:attack}, our method maintains consistent performance across prompt, structural, and combined attacks, indicating robustness to multiple sources of perturbation.

\begin{table}
    \centering
    \caption{GNN backbone for different components.}
    \begin{tabular}{lcc}
        \Xhline{1.0pt}
        \textbf{Variants} & \textbf{MMLU} & \textbf{MultiArith }   \\
        \Xhline{1.0pt}
        Ordering (RGAT), Denoising (GAT)  & 82.35 & 97.97 \\
        Ordering (RGAT), Denoising (GCN)   & 82.30  &  98.14   \\
        Ordering (RGCN), Denoising (GCN)  & 82.23 &  98.64   \\
        Ordering (RGCN), Denoising (GAT)  & 83.66 &   98.81 \\
        \Xhline{1.0pt}
    \end{tabular}
    \label{tab:backbone}
\end{table}

\begin{table}
    \centering
    \caption{Model performance under different attacks.}
    \begin{tabular}{lccc}
        \Xhline{1.0pt}
        \textbf{Variants} & \textbf{MMLU} & \textbf{GSM8K} & \textbf{MultiArith }   \\
        \Xhline{1.0pt}
        Prompt Attack  & 83.14 & 92.12 & 98.13 \\
        Structure Attack   & 82.08 & 91.82 &  98.64   \\
        Prompt \& Structure Attack & 81.88 & 91.46 &  97.79   \\
        RADAR  & 83.66 & 92.51 &   98.81 \\
        \Xhline{1.0pt}
    \end{tabular}
    \label{tab:attack}
\end{table}

\section{Case Study}
\begin{figure}
    \centering
    \begin{subfigure}{0.3\linewidth}
        \includegraphics[width=\linewidth]{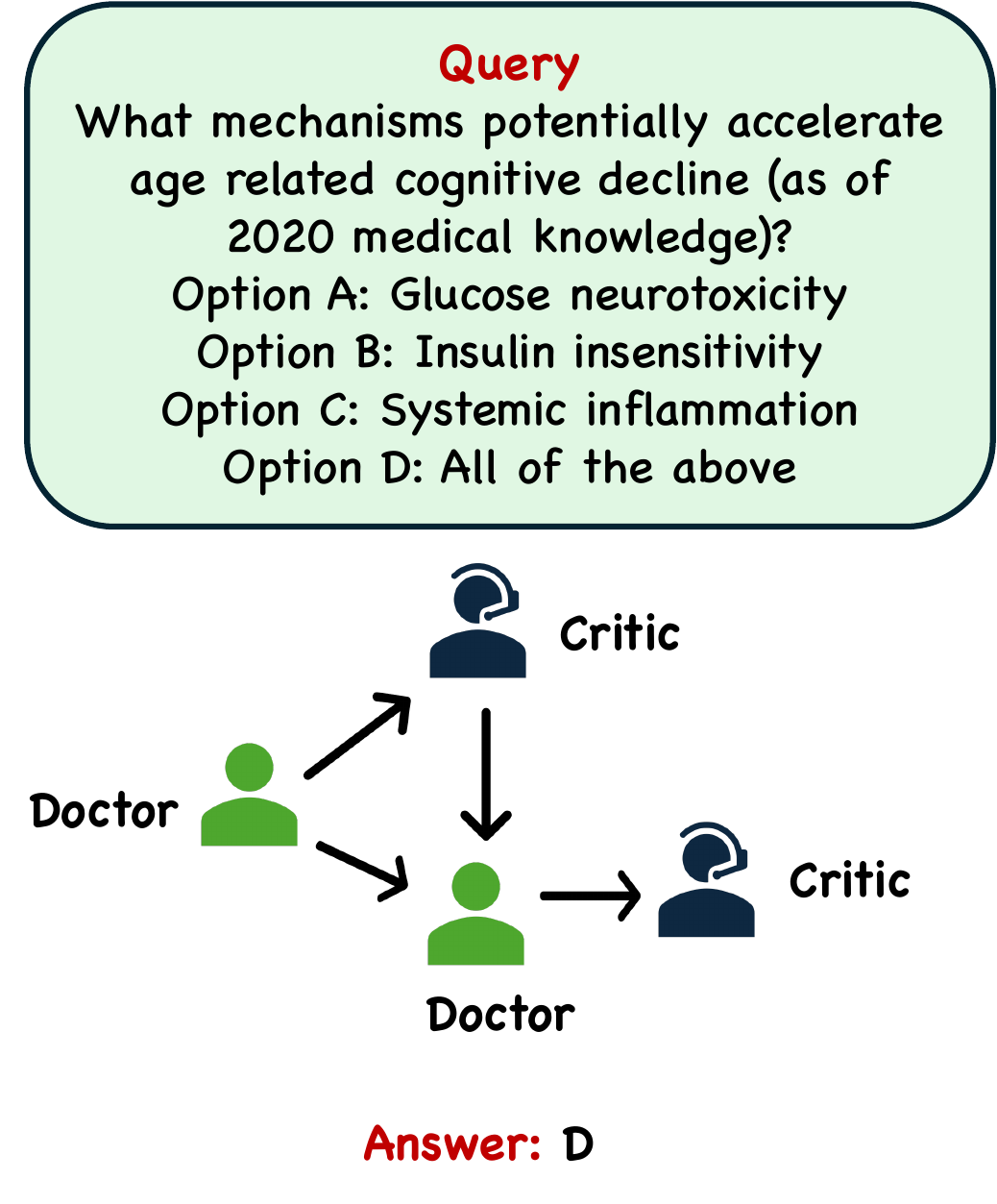} 
        \caption{Sampled case in MMLU.}
    \end{subfigure}
    \begin{subfigure}{0.3\linewidth}
        \raisebox{0.1\height}{
        \includegraphics[width=\linewidth]{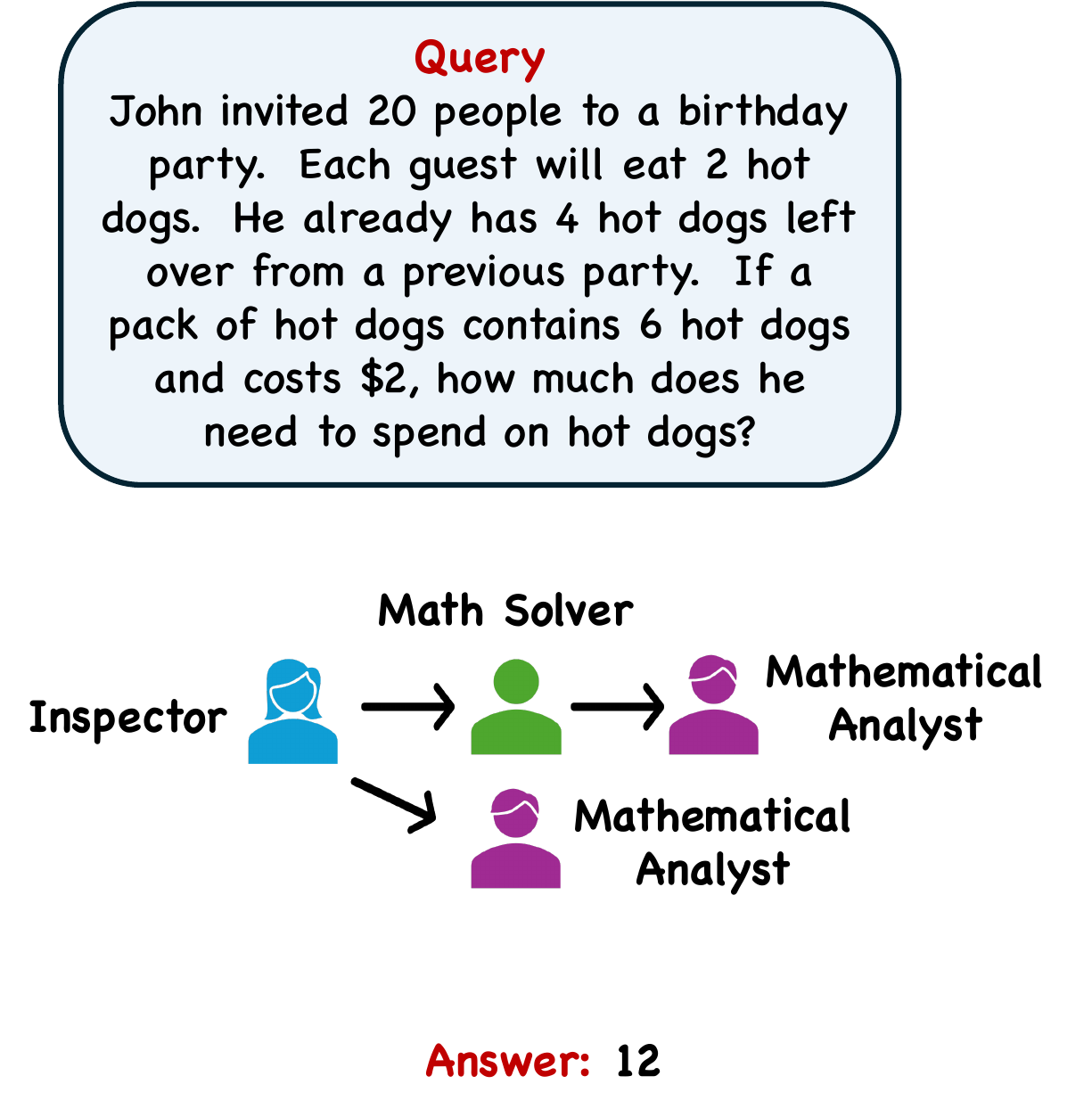} 
        }
        \caption{Sampled case in GSM8K.}
    \end{subfigure}
    \begin{subfigure}{0.3\linewidth}
        \includegraphics[width=\linewidth]{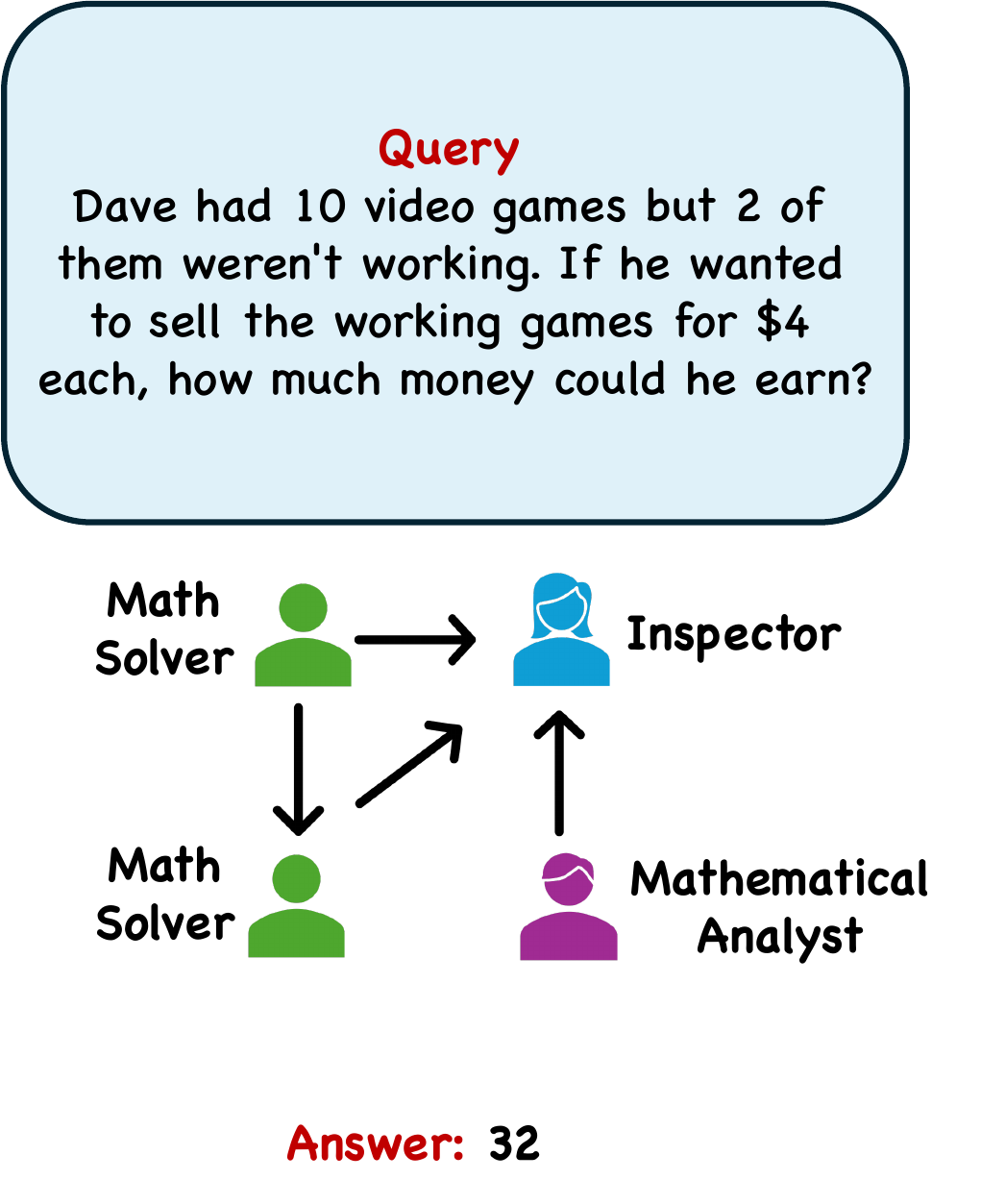} 
        \caption{Sampled case in MultiArith.}
    \end{subfigure}    
    \begin{subfigure}{0.3\linewidth}
        \raisebox{0.2\height}{
        \includegraphics[width=\linewidth]{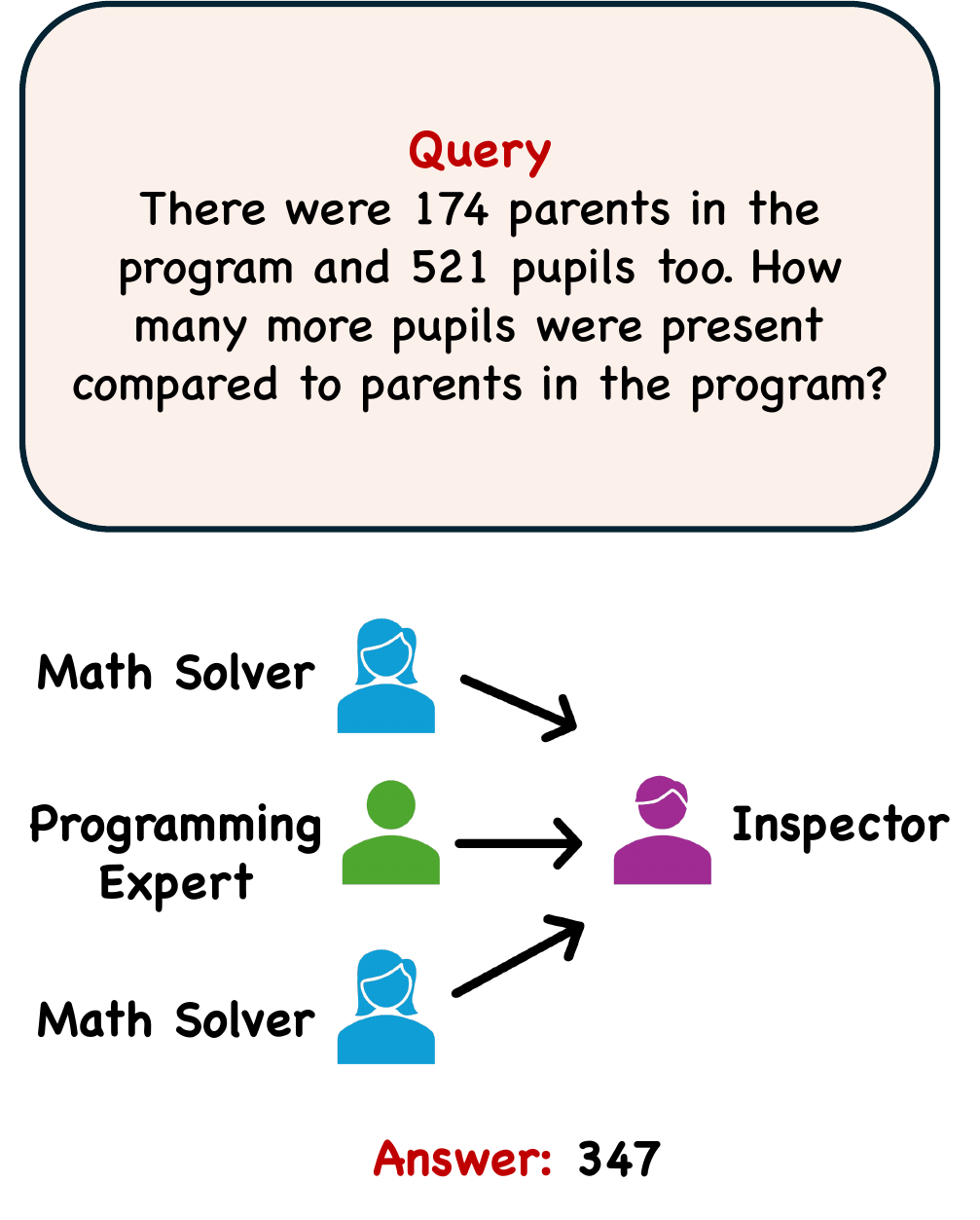} 
        }
        \caption{Sampled case in SVAMP.}
    \end{subfigure}
    \begin{subfigure}{0.3\linewidth}
        \raisebox{0.4\height}{
        \includegraphics[width=\linewidth]{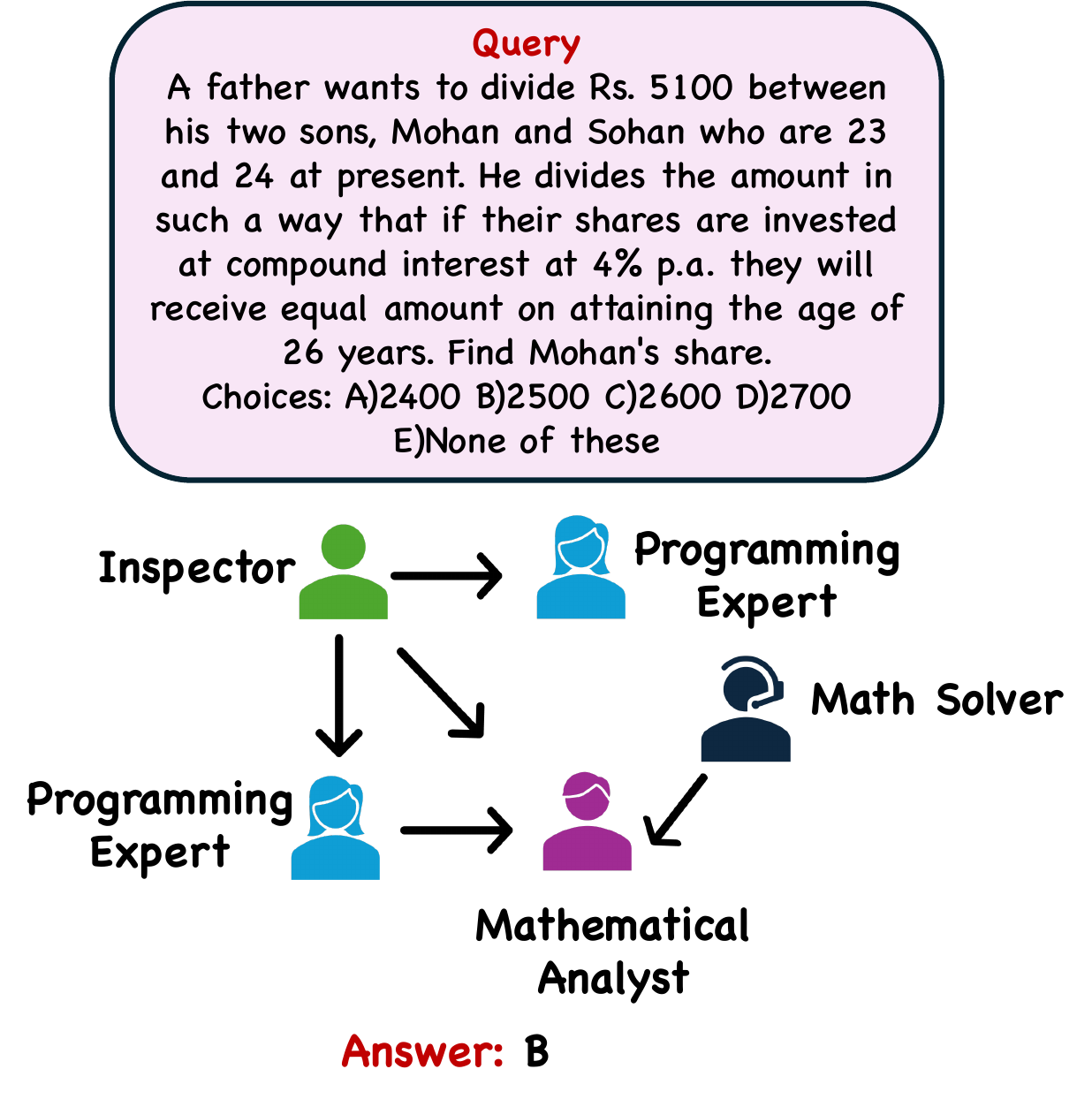} 
        }
        \caption{Sampled case in AQuA.}
    \end{subfigure}
    \begin{subfigure}{0.3\linewidth}
        \includegraphics[width=\linewidth]{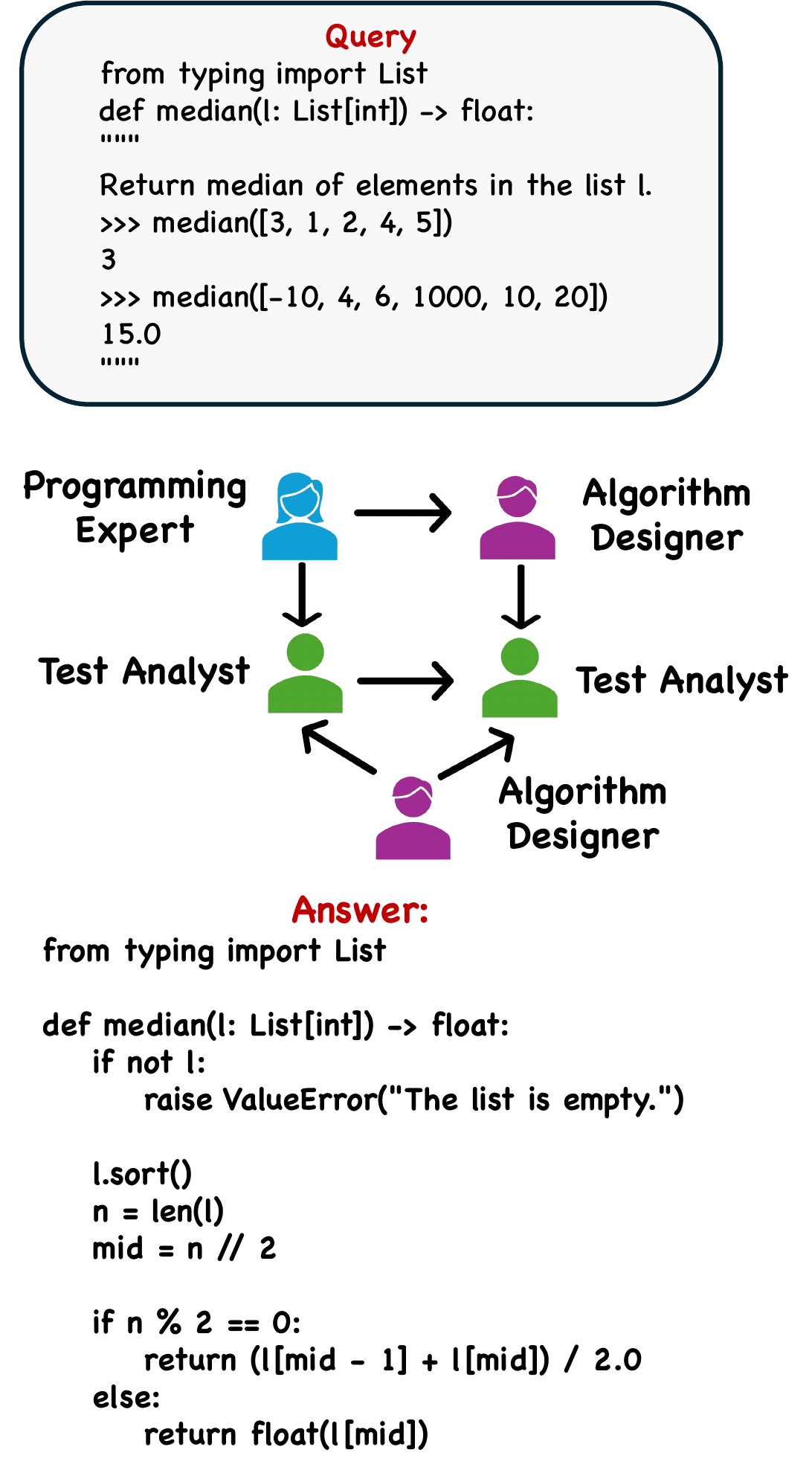} 
        \caption{Sampled case in HumanEval.}
    \end{subfigure}
    \caption{Case study of the communication topologies generated by RADAR.}
    \label{fig:ab_case}
\end{figure}

Figure \ref{fig:ab_case} illustrates the communication topologies generated by RADAR. As we can see, the resulting structures are task-adaptive and vary across different tasks and datasets. Although the maximum number of agents is fixed at five, RADAR does not necessarily activate all agents for simpler tasks. In contrast, for more complex benchmarks such as AQUA and HumanEval, all five agents are utilized in the generated topology, reflecting the increased reasoning and coordination requirements of these tasks.

%%%%%%%%%%%%%%%%%%%%%%%%%%%%%%%%%%%%%%%%%%%%%%%%%%%%%%%%%%%%%%%%%%%%%%%%%%%%%%%
%%%%%%%%%%%%%%%%%%%%%%%%%%%%%%%%%%%%%%%%%%%%%%%%%%%%%%%%%%%%%%%%%%%%%%%%%%%%%%%

\end{document}